%% file: main.tex
\documentclass{article} %
\usepackage{iclr2027_conference,times}

\input{math_commands.tex}

\usepackage{hyperref}
\hypersetup{hidelinks}
\usepackage{url}

\usepackage{booktabs}   %
\usepackage{multirow}   %
\usepackage{graphicx}   %
\usepackage{makecell}   %
\usepackage{xspace}
\usepackage{amsmath}
\usepackage{wrapfig}
\usepackage{subcaption}
\usepackage[many]{tcolorbox}

\iclrfinalcopy %

\title{DRIFT: Data Selection for LLM Instruction Tuning via On-Policy Attribution}

\newcommand{\mymethod}{DRIFT\xspace}

\author{%
  Zefan Wang \quad
  Lincheng Li \quad
  Tianyu Yu \quad
  Yuan Yao\thanks{Corresponding author.} \\
  \vspace{0.1cm}
  Tsinghua University \\
  \texttt{zefan.wang.thu@gmail.com} \quad \texttt{yaoyuanthu@gmail.com} %
}

\newtheorem{theorem}{Theorem}
\newtheorem{proposition}[theorem]{Proposition}
\newtheorem{corollary}[theorem]{Corollary}

\begin{document}

\maketitle
\fancyhead{} %
\renewcommand{\headrulewidth}{0pt}

\begin{abstract}

Data selection for instruction tuning is usually studied for efficiency, reaching strong performance with a fraction of the training data. We ask whether data selection can instead raise the performance ceiling, producing a better model than training on all the available data. 
To test this, we take a large language model (LLM) that has already completed supervised fine-tuning (SFT) on its full corpus and train it further on examples selected from that same corpus. Any gain therefore lifts the model above the ceiling set by full-data training. 
Existing selection methods yield little or no gain in this setting. 
We propose DRIFT (Data Refinement via On-Policy Influence Functions for Supervised Fine-Tuning), a data attribution method built on influence functions. DRIFT samples the model's own responses to validation queries and weights them by correctness to form a validation objective. 
It ranks the examples of the original corpus by their estimated influence on this objective, and the model then continues SFT on the top-ranked subset. This design is motivated by the \emph{proximity gap} in influence function theory: we hypothesize that these on-policy responses are better validation targets than external references in terms of locality. DRIFT also corrects influence scores for their dependence on gradient norm within each validation task. DRIFT outperforms all evaluated data selection baselines on Olmo3-7B-Instruct-SFT and OpenR1-Distill-7B, raising average accuracy across eight benchmarks from 39.17 to 40.14 and from 57.23 to 58.66, respectively, with gains on benchmarks not used for attribution.
\end{abstract}

\input{chapters/introduction.tex}
\input{chapters/preliminary.tex}
\input{chapters/method.tex}
\input{chapters/experiment.tex}
\input{chapters/discussion.tex}
\input{chapters/relatedworks.tex}
\input{chapters/conclusion.tex}

\clearpage
\section*{Reproducibility Statement}
We will release the attribution code, identifiers for the selected top-10\% subset, validation rollouts with their correctness labels, and per-seed evaluation results. Appendix~\ref{app:setup} specifies the training and evaluation settings, Table~\ref{tab:attr_hparams} lists the attribution-stage hyperparameters, and Appendix~\ref{app:cost} reports compute costs.

\section*{AI Use Statement}
We used LLM-based assistants to generate reviewer-style critiques that informed additional experiments and helped align the scope of our claims with the supporting evidence. These tools also assisted with language editing, formatting, and manuscript organization. The authors evaluated the feedback and retain full responsibility for the research, the final manuscript, and all reported results.

\bibliography{reference}
\bibliographystyle{iclr2027_conference}

\appendix
\input{chapters/appendix.tex}

\end{document}

%% file: math_commands.tex
\usepackage{amsmath,amsfonts,bm}

\def\eqref#1{equation~\ref{#1}}

\def\1{\bm{1}}

\DeclareMathAlphabet{\mathsfit}{\encodingdefault}{\sfdefault}{m}{sl}
\SetMathAlphabet{\mathsfit}{bold}{\encodingdefault}{\sfdefault}{bx}{n}

%% file: chapters/introduction.tex
\section{Introduction}
\label{sec:introduction}

Supervised fine-tuning (SFT) trains large language models (LLMs) to follow instructions, and the distribution of its training data shapes the capabilities they acquire \citep{ouyang2022training,wei2021finetuned,zhou2023lima,ye2024data}. Data selection for instruction tuning is usually studied as a question of efficiency. Selected subsets are compared with random subsets under a restricted training budget, and the goal is to reach strong performance with a fraction of the data \citep{wettig2024qurating,xia2024less,ivison2025large,xie2023data}. This leaves open a more ambitious question. \emph{Can data selection raise the performance ceiling, producing a better model than training on all the available data?}

We study this question in a direct setting. We start from an LLM that has completed SFT on its full corpus, select examples from the same corpus, and continue training on them under a fixed budget and schedule. The fully trained model marks the ceiling of full-data training. It has already seen every candidate example, so any further gain has to come from a better distribution over data it already has. Such gains are hard to obtain. In our main experiments, most existing selection methods fail to improve the fully trained model~(Table~\ref{tab:merged_models}), and several make it worse.

Instance-level data attribution gives finer control over this choice than reweighting broad data categories \citep{xie2023doremi,liu2024regmix,peng2020domain2vec}. Influence functions (IF) \citep{koh2017understanding,grosse2023studying} estimate how up-weighting a training example changes the loss on a validation set, and data selection methods based on IF keep the examples predicted to reduce that loss the most \citep{pruthi2020estimating,ilyas2022datamodels}. What IF selects therefore depends on the validation targets. The common choice is cross-entropy (CE) on reference responses, the known correct answers to the validation queries \citep{xia2024less,wang2024greats}.
There is a mismatch between reference responses and how IF works, as IF is a local estimate. In neural networks, IF tracks retraining that stays close to the current checkpoint, and its discrepancy from unconstrained retraining is termed the \textbf{proximity gap} \citep{bae2022if} (Section~\ref{sec:fragility}). A reference response may use phrasing or reasoning steps that the model rarely produces. Fitting it may require moving away from the checkpoint, which is where the influence estimate is least informative.

A direct way to avoid this mismatch is to take validation targets from the model's on-policy responses, as suggested by reinforcement learning research. 
\citep{mukherjee2025reinforcement} report that RL leaves most of a model's parameters effectively unchanged, and conjecture that this is mainly because the training data lie near the model's own distribution. Following this observation, we hypothesize that validation targets sampled from the model keep the attribution objective within the neighborhood where influence estimates are informative.

We propose \textbf{DRIFT} (\textbf{D}ata \textbf{R}efinement via On-Policy \textbf{I}nfluence \textbf{F}unctions for Supervised Fine-\textbf{T}uning) based on this hypothesis. DRIFT samples responses to the validation queries from the checkpoint and checks their correctness. Correct responses receive weight +1 and incorrect ones -1. Weighting their CE gradients this way gives the gradient of a policy-gradient (PG) validation loss, which states which of the model's current behaviors should become more likely. IF with this loss ranks every example in the original corpus, and the model continues ordinary SFT on the top-ranked subset. The generated responses define the selection objective, while the training data still come from the original corpus.

A second issue is \emph{gradient norm bias}. An influence score is an inner product, so an example with a large gradient norm can rank high even when its gradient is poorly aligned with the validation objective, as happens with some malformed responses (Appendix~\ref{app:qualitative_cases}). DRIFT fits the relationship between score magnitude and gradient norm in log space for each validation task and removes the fitted norm component.

On Olmo3-7B-Instruct-SFT, DRIFT raises the average accuracy over eight benchmarks from 39.17\% to 40.14\%. In contrast, no evaluated baseline improves the starting checkpoint by more than 0.4 points, and several fall below it. The gains extend to four benchmarks whose queries are never used for attribution, and they hold across training seeds and repeated selection runs. On the reasoning-focused OpenR1-Distill-7B, DRIFT improves the average from 57.23\% to 58.66\%. Ablations show that both the signed on-policy objective and influence-based scoring contribute to the gain. Together, these results show that a model that has completed its SFT can still improve by revisiting its own training data under a better distribution.

%% file: chapters/preliminary.tex
\section{Preliminaries}
\label{sec:preliminaries}

\subsection{Influence Functions and Practical Approximations for LLMs}
\label{sec:if_approximations}

Influence functions estimate how up-weighting a training example $z_{\mathrm{train}}$ changes the loss on a validation target $z_{\mathrm{val}}$, without retraining the model. Let $\theta^\ast$ denote the parameters of a model trained to minimize the empirical risk $\frac{1}{n}\sum_{i=1}^{n}\mathcal{L}(z_i;\theta)$. The first-order effect of an infinitesimal increase in the weight of $z_{\mathrm{train}}$ is

\begin{equation}
\mathcal{I}(z_{\mathrm{train}},z_{\mathrm{val}})
= -\nabla_\theta\mathcal{L}(z_{\mathrm{val}};\theta^\ast)^\top
H_{\theta^\ast}^{-1}
\nabla_\theta\mathcal{L}(z_{\mathrm{train}};\theta^\ast),
\label{eq:standard_if}
\end{equation}

where $H_{\theta^\ast}=\frac{1}{n}\sum_{i=1}^{n}\nabla_\theta^2\mathcal{L}(z_i;\theta^\ast)$ is the Hessian of the empirical risk.

For LLMs, practical estimators either approximate the inverse Hessian \citep{grosse2023studying,choe2024your} or replace it with the identity, reducing the estimate to a gradient inner product \citep{zhu2025data,xia2024less}. Gradient sparsification and random projection make it feasible to store features for an entire training corpus \citep{park2023trak,hu2025grass}. Specifically, a sparse matrix $P\in\mathbb{R}^{d\times D}$ maps the full gradient from $\mathbb{R}^{D}$ to $\mathbb{R}^{d}$, with $d\ll D$. The Johnson--Lindenstrauss lemma motivates this approximation: suitable projections approximately preserve pairwise geometry when the projected dimension is sufficiently large \citep{william1984extensions}.

Writing the projected gradient as $\phi(z;\theta^\ast)=P\nabla_\theta\mathcal{L}(z;\theta^\ast)$, we obtain the approximate score

\begin{equation}
S(z_{\mathrm{train}},z_{\mathrm{val}})
:= -\mathcal{I}(z_{\mathrm{train}},z_{\mathrm{val}})
\approx \phi(z_{\mathrm{train}};\theta^\ast)^\top
\phi(z_{\mathrm{val}};\theta^\ast).
\label{eq:projected_if}
\end{equation}

With this sign convention, a positive score means that up-weighting the example reduces the validation loss.

\paragraph{Gradient norm bias.}
An inner product reflects both direction and magnitude. Examples with large gradient norms can therefore dominate the ranking, including short responses \citep{xia2024less} and outliers \citep{hu2025grass}. Section~\ref{sec:analysis} compares several corrections.

\subsection{The Fragility of Influence Functions}
\label{sec:fragility}

\subsubsection{The Non-Convergence Gap}
\label{sec:non_convergence}

Classical IF assumes a stationary checkpoint. Away from stationarity, further optimization moves the model even without a change to the data, potentially obscuring the effect of the data intervention \citep{bae2022if}. This \textbf{non-convergence gap} motivates computing attribution after the original SFT schedule. We treat the completed checkpoint as approximately stationary; Appendix~\ref{app:non_convergence} gives the local quadratic formulation.

\subsubsection{The Proximity Gap and On-Policy Attribution}
\label{sec:proximity}

External reference responses provide a straightforward validation objective: reduce CE on known correct answers \citep{xia2024less,wang2024greats}. These targets are \textbf{off-policy}, however, and may use phrasing or reasoning paths that the model rarely produces. The issue for influence-based selection is whether the changes needed to fit them are captured by a local approximation around the checkpoint.

The \textbf{proximity gap} makes this limitation explicit. In neural networks, practical IF estimates closely track the Proximal Bregman Response Function (PBRF), which encourages outputs and parameters to remain close to the current model \citep{bae2022if}. The gap measures the difference between warm-start retraining optima with and without a parameter-proximity restriction.
Reference-based CE does not account for this restriction.

\textbf{On-policy responses} offer an alternative tied to the model's current behavior. Under exact model sampling, their full-sequence log-likelihood gradients cancel in expectation. Nonuniform weights instead express preferences among these responses, as formalized below.

\begin{proposition}[On-Policy Validation Gradients]
\label{thm:on_policy_shift}
Fix a query $x$ and a checkpoint $\theta^\ast$, and let $w(y)$ be a bounded response weight that does not depend on $\theta$. For responses sampled as $y\sim\pi_{\theta^\ast}(\cdot\mid x)$, define

\[
s(y)=\nabla_\theta\log\pi_{\theta^\ast}(y\mid x),
\qquad
 g_w=\mathbb{E}[w(y)s(y)],
\qquad
G_x^2=\mathbb{E}\|s(y)\|_2^2<\infty,
\]

where expectations are under $\pi_{\theta^\ast}(\cdot\mid x)$.

\noindent\textbf{Expected gradient.}
\(
g_w=\left.\nabla_\theta\mathbb{E}_{y\sim\pi_\theta(\cdot\mid x)}[w(y)]\right|_{\theta^\ast}.
\)
In particular, constant weights give a zero expected full-sequence CE gradient.

\noindent\textbf{Validation-step bound.} For $\Delta\theta_{\mathrm{val}}=\eta g_w$, with $\eta\geq 0$,
\(
\|\Delta\theta_{\mathrm{val}}\|_2\leq\eta G_x\sqrt{\operatorname{Var}_{\pi_{\theta^\ast}}[w(y)]}.
\)

\end{proposition}

Proposition~\ref{thm:on_policy_shift} concerns a single population-level validation step at a fixed checkpoint, and its proof appears in Appendix~\ref{sec:appendix_proof}. Within this scope, when the weights encode response quality, the proposition connects the validation gradient to improving expected quality under the current model. 
As an empirical check of the locality hypothesis during training, 
Figure~\ref{fig:proximity}
tracks parameter drift under continued SFT. After the initial steps, on-policy selections change the parameters less than reference-based selection.

%% file: chapters/method.tex
\section{DRIFT: Data Refinement via On-Policy Influence Functions for Supervised Fine-Tuning}
\label{sec:method}

DRIFT refines the SFT data distribution through on-policy influence scoring. Starting from a checkpoint that has completed its original SFT schedule, it uses the model's responses to identify examples worth revisiting in the original corpus. Figure~\ref{fig:main} summarizes the four steps.

\begin{figure}[!t]
    \centering
    \input{fig/draft_reward_labels}
    \caption{Overview of DRIFT. On-policy rollouts and correctness rewards define task-specific PG validation centroids for scoring the original SFT corpus. After correcting gradient norm bias, we continue SFT on the highest-ranked examples.}
    \label{fig:main}
    \vspace{-6pt}
\end{figure}

\paragraph{Step 1: On-Policy Proximal Validation Construction.}
Given validation queries $X_{\mathrm{val}}$, we sample $N_x$ responses to each query $x$ from the base model $\theta^\ast$ with temperature and top-p sampling (Appendix~\ref{app:detailed_setup}). The resulting rollouts replace external references as validation targets.

\paragraph{Step 2: Policy Gradient IF Formulation.}
To use both successful and failed rollouts, we replace the reference-based CE validation objective with a PG objective. Each rollout receives a reward $r^{(i)}=r(y^{(i)})\in\{+1,-1\}$ from the task's evaluator, $+1$ if it is correct and $-1$ otherwise. Using the token-mean CE loss of SFT, we define the PG objective for query $x$ as
\begin{equation}
\mathcal{J}_{PG}(\theta;x)=-\frac{1}{N_x}\sum_{i=1}^{N_x} r^{(i)}\,\mathcal{L}_{CE}\big(x,y^{(i)};\theta\big),
\qquad
\mathcal{L}_{CE}(x,y;\theta)=-\frac{\log\pi_\theta(y\mid x)}{T(y)},
\label{eq:pg_if}
\end{equation}
where $T(y)$ counts the assistant tokens in $y$. Because the rollouts come from temperature and top-p sampling and the objective is averaged over tokens, $\mathcal{J}_{PG}$ is a surrogate of expected reward. Under exact sampling and a full-sequence objective, its expected gradient equals the gradient of expected reward (Proposition~\ref{thm:on_policy_shift} with $w=r$, Appendix~\ref{sec:appendix_proof}).
The gradient $\nabla_\theta\mathcal{J}_{PG}$ at $\theta^\ast$ is a correctness-weighted combination of the per-rollout gradients $\nabla_\theta\mathcal{L}_{CE}(x,y^{(i)};\theta^\ast)$. Each of them serves as a \textbf{validation anchor}, computed exactly like the gradient of a corpus example.

To balance tasks, we average over rollouts within each query and over queries within each task, so each task carries equal total absolute weight. A rollout $z_{\mathrm{val}}=(x,y)$ of query $x$ in task $D_k$ thus receives the coefficient $\alpha(z_{\mathrm{val}})=r(y)/(|D_k|N_x)$, where $|D_k|$ is the number of queries in task $k$. We use GraSS \citep{hu2025grass} to obtain sparse projections $\phi(z;\theta^\ast)\in\mathbb{R}^d$ of $\nabla_\theta\mathcal{L}_{CE}$ for both anchors and corpus examples.

Summing the projected anchors with these coefficients gives a \textbf{validation centroid} for each task. The raw score of a training example is its dot product with this centroid,
\begin{equation}
S_{\mathrm{raw}}^{(k)}(z_{\mathrm{train}})
=\phi(z_{\mathrm{train}};\theta^\ast)^\top
\Big(\sum_{z_{\mathrm{val}}\in\mathcal{R}_k}\alpha(z_{\mathrm{val}})\,\phi(z_{\mathrm{val}};\theta^\ast)\Big),
\label{eq:raw_score}
\end{equation}
where $\mathcal{R}_k$ contains the rollouts from $D_k$. A positive score predicts that training on $z_{\mathrm{train}}$ increases $\mathcal{J}_{PG}$.

\paragraph{Step 3: Influence Debiasing.}
The raw score depends on both the norm of the training gradient and its alignment with the centroid. Cosine normalization removes the explicit norm factor, but normalized scores can remain associated with gradient norm when norm and direction vary together across examples. We therefore model this association with a power law, $|S|\propto\|\phi\|^\beta$, fitted separately for each task. For task $k$, we regress $\log|S_{\mathrm{raw}}^{(k)}(z_{\mathrm{train}})|$ on $\log\|\phi(z_{\mathrm{train}};\theta^\ast)\|$ with an intercept to obtain $\beta^{(k)}$, and then remove the fitted log-norm component while keeping the sign of the score,
\begin{equation}
\widetilde{S}^{(k)}(z_{\mathrm{train}})
=\operatorname{sgn}\!\left(S_{\mathrm{raw}}^{(k)}(z_{\mathrm{train}})\right)
\exp\!\left(
\log\left|S_{\mathrm{raw}}^{(k)}(z_{\mathrm{train}})\right|
-\beta^{(k)}\log\|\phi(z_{\mathrm{train}};\theta^\ast)\|
\right).
\label{eq:debiased_score}
\end{equation}
The overall score is $S(z_{\mathrm{train}})=\sum_k\widetilde{S}^{(k)}(z_{\mathrm{train}})$. Appendix~\ref{app:method_details} discusses the fitted slopes and the statistical interpretation.

\paragraph{Step 4: Continual Training with Refined Distribution.}
We rank the candidate examples by $S(z_{\mathrm{train}})$, select the top 10\% by default, and continue SFT on this subset from $\theta^\ast$.

%% file: fig/draft_reward_labels.tex
\begingroup
\definecolor{rewardbackground}{RGB}{218,227,243}
\begin{tikzpicture}
\node[anchor=south west,inner sep=0] (artwork) at (0,0)
  {\includegraphics[width=\textwidth]{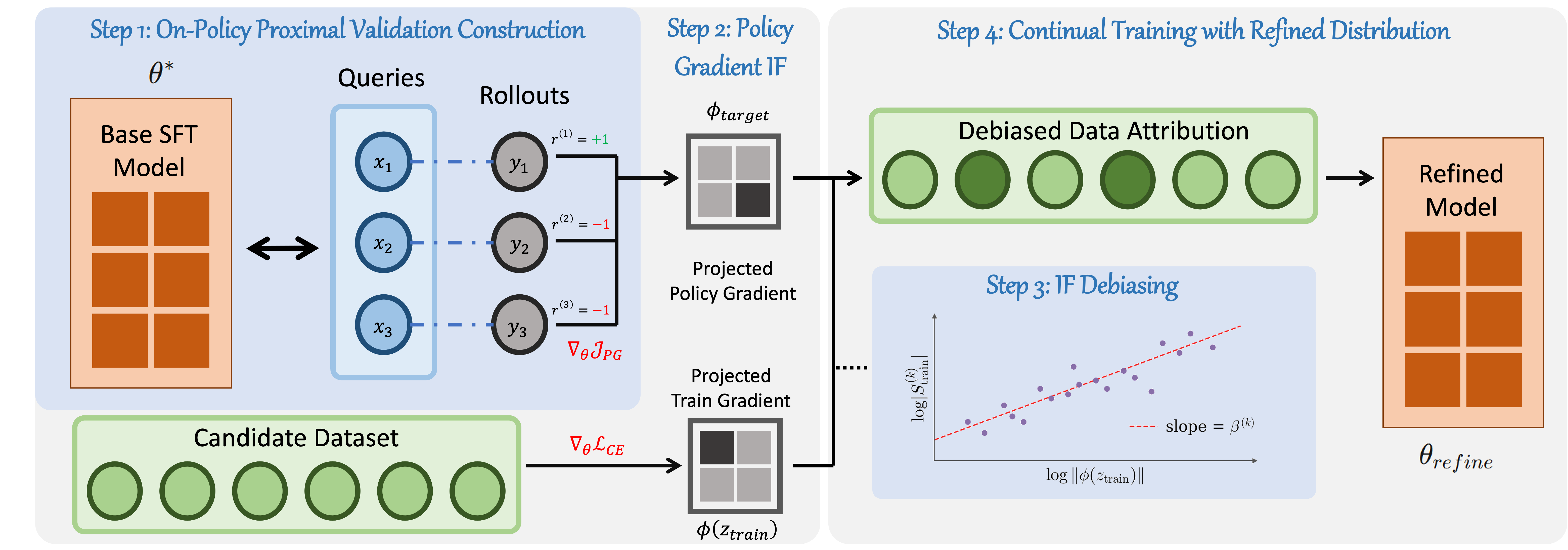}};
\end{tikzpicture}
\endgroup

%% file: chapters/experiment.tex
\section{Experiments}
\label{sec:experiments}

\subsection{Experimental Setup}
\label{sec:experimental_setup}

\paragraph{Base Model and Training Objective.}
Our main experiments use Olmo3-7B-Instruct-SFT \citep{olmo2025olmo}, a general-purpose instruction model with publicly available SFT data and a training recipe. The candidate corpus is Dolci-Instruct-SFT, which contains approximately 2M examples. Starting from the checkpoint after its original SFT schedule, each corpus selection method selects 10\% of these examples for continual training. We hold the selection ratio and training schedule fixed and compare improvement over the starting checkpoint. Appendix~\ref{app:setup} gives dataset statistics and hyperparameters. We also apply the same protocol to the reasoning-focused OpenR1-Distill-7B model \citep{openr1}; Table~\ref{tab:merged_models} reports its aggregate results, and Appendix~\ref{app:openr1} gives the settings and per-benchmark results.

\paragraph{Validation Set Construction.}
The validation set covers math, code, reasoning, and knowledge through MATH \citep{hendrycks2021math}, MBPP, BBH \citep{suzgun2022challenging}, and MMLU-Pro \citep{wang2024mmlu}. The MBPP queries are original problems disjoint from the MBPP+ evaluation set \citep{evalplus}. Query counts and task-selection criteria are given in Appendix~\ref{app:setup}. DRIFT generates responses to these queries and verifies their correctness; the other validation-dependent corpus selectors in Table~\ref{tab:merged_models} use the original reference responses.

\paragraph{Evaluation Benchmarks.}
We evaluate on eight benchmarks spanning the same four capability areas. To test whether gains extend beyond the benchmark sources used for attribution, we report two groups. The \textbf{validation-source benchmarks} are BBH \citep{suzgun2022challenging}, MBPP+ \citep{evalplus}, MATH \citep{hendrycks2021math}, and MMLU-Pro \citep{wang2024mmlu}. The \textbf{additional benchmarks} are ZebraLogic \citep{zebralogic2024}, LiveCodeBench v5 \citep{jain2024livecodebench}, OlympiadBench \citep{he2024olympiadbench}, and GPQA Diamond \citep{rein2023gpqa}. No queries from the additional benchmarks are used for validation or attribution. Both groups cover math, code, reasoning, and knowledge. \textbf{All} denotes the mean accuracy across all eight benchmarks; \textbf{Additional} denotes the mean across the four additional benchmarks.

\paragraph{Baselines.}
All corpus selectors use the same 10\% selection budget. We compare DRIFT with uniform Random selection; distribution-matching DSIR \citep{xie2023data}; representation-based RDS \citep{ivison2025large}; quality-based Qurating \citep{wettig2024qurating}; lexical BM25 \citep{robertson2009probabilistic}; and the influence-based methods LESS \citep{xia2024less} and IF (GraSS) \citep{hu2025grass}. Appendices~\ref{app:setup} and \ref{app:baselines} give the evaluation settings and baseline implementations.

\input{table/main_results}
\subsection{Main Results}
\label{sec:main_results}

\paragraph{DRIFT improves the starting checkpoint and outperforms the evaluated baselines.}
On Olmo3-7B-Instruct-SFT, DRIFT raises the All average from 39.17 to 40.14, exceeding BM25 by 0.60 points and Random by 1.10 points (Table~\ref{tab:merged_models}). It also achieves the highest Additional average. The Olmo3 continual-training results are means over three seeds. Most evaluated selectors, including DSIR, LESS, and RDS, do not improve All over the starting checkpoint. This distinction is central to refinement: outperforming Random is not enough if the refined model still falls below its starting point.

\paragraph{DRIFT also improves the reasoning-focused model.}
On OpenR1-Distill-7B, DRIFT raises All from 57.23 to 58.66 and Additional from 46.24 to 48.73 (Table~\ref{tab:merged_models}). Reference-based IF is the strongest baseline on this model, yet DRIFT exceeds it by 0.81 points on All. Appendix~\ref{app:openr1} gives the full results. 

\paragraph{Robustness to selection-stage randomness.}
Across rollout seeds and projection seeds, the top-10\% selections have Jaccard overlaps of 0.73 and 0.69, respectively. Three additional downstream runs achieve All scores of 40.03, 40.19, and 40.08, each above the strongest baseline's average of 39.54 (Appendix~\ref{app:robustness}).

\paragraph{Attribution on the original SFT corpus outperforms self-distillation.}
In these experiments, direct training on verified rollouts reduces All to 37.64 on Olmo3 and 49.33 on OpenR1, compared with 40.14 and 58.66 for DRIFT. Mixing the rollouts with random corpus examples on Olmo3 recovers part of the loss (38.37), but still falls below the starting checkpoint. Appendices~\ref{app:self_distillation_results} and \ref{app:openr1} give the full results.

%% file: table/main_results.tex
\begin{table}[t]
\centering
\caption{Continual-SFT accuracy (\%) on Olmo3-7B-Instruct-SFT (three-seed means) and OpenR1-Distill-7B, using 10\% of each model's original SFT corpus. Base denotes the starting checkpoint; Rand., random selection; Add., the additional-benchmark average (Section~\ref{sec:experiments}). Bold marks the best continual-training result per column; colored differences are relative to Base. Full results and self-distillation comparisons appear in Appendices~\ref{app:robustness} and \ref{app:openr1}.}
\label{tab:merged_models}
\begingroup
\fontsize{9}{10.7}\selectfont
\setlength{\tabcolsep}{2.5pt}
\renewcommand{\arraystretch}{1.12}
\resizebox{\linewidth}{!}{%
\begin{tabular}{@{}l|cccc|cccc|cc|cc@{}}
\toprule
 & \multicolumn{10}{c|}{\textbf{Olmo3-7B-Instruct-SFT}} & \multicolumn{2}{c}{\textbf{OpenR1-Distill-7B}} \\
\cmidrule(lr){2-11} \cmidrule(l){12-13}
 & \multicolumn{4}{c|}{\textbf{Validation-Source Benchmarks}} & \multicolumn{4}{c|}{\textbf{Additional Benchmarks}} & \multicolumn{2}{c|}{\textbf{Average}} & \multicolumn{2}{c}{\textbf{Average}} \\
\cmidrule(lr){2-5} \cmidrule(lr){6-9} \cmidrule(lr){10-11} \cmidrule(l){12-13}
\makecell[l]{\textbf{Method}\\Acc. (\%)} & \makecell{\textbf{BBH}\\} & \makecell{\textbf{MBPP+}\\} & \makecell{\textbf{MATH}\\} & \makecell{\textbf{MMLU}\\\textbf{Pro}} & \makecell{\textbf{Zebra}\\\textbf{Logic}} & \makecell{\textbf{LCB}\\\textbf{v5}} & \makecell{\textbf{Olympiad}\\\textbf{Bench}} & \makecell{\textbf{GPQA}\\\textbf{Diamond}} & \makecell{\textbf{All}\\} & \makecell{\textbf{Add.}\\} & \makecell{\textbf{All}\\} & \makecell{\textbf{Add.}\\} \\
\midrule
\textbf{Base} & 44.63 & 56.42 & 66.16 & 44.22 & 17.23 & 15.43 & 34.27 & 35.01 & 39.17 & 25.49 & 57.23 & 46.24 \\
\midrule
Rand. & 44.45 & 56.34 & 66.40 & 44.25 & 17.67 & 14.38 & 35.04 & 33.77 & 39.04 \textcolor[rgb]{0.89, 0.65, 0.65}{\scriptsize -0.13} & 25.21 \textcolor[rgb]{0.87, 0.60, 0.60}{\scriptsize -0.28} & 57.40 \textcolor[rgb]{0.62, 0.85, 0.62}{\scriptsize +0.17} & 46.65 \textcolor[rgb]{0.59, 0.84, 0.59}{\scriptsize +0.41} \\
DSIR & 39.97 & 52.63 & 55.11 & 36.22 & 17.68 & 12.73 & 27.83 & 32.82 & 34.37 \textcolor[rgb]{0.70, 0.00, 0.00}{\scriptsize -4.80} & 22.76 \textcolor[rgb]{0.70, 0.00, 0.00}{\scriptsize -2.73} & 56.24 \textcolor[rgb]{0.87, 0.61, 0.61}{\scriptsize -0.99} & 45.43 \textcolor[rgb]{0.88, 0.64, 0.64}{\scriptsize -0.81} \\
Qurating & 44.27 & 55.96 & 66.89 & 44.14 & 17.77 & 15.49 & 34.98 & 34.61 & 39.26 \textcolor[rgb]{0.67, 0.88, 0.67}{\scriptsize +0.09} & 25.71 \textcolor[rgb]{0.62, 0.86, 0.62}{\scriptsize +0.22} & 57.78 \textcolor[rgb]{0.43, 0.75, 0.43}{\scriptsize +0.55} & 47.05 \textcolor[rgb]{0.47, 0.77, 0.47}{\scriptsize +0.81} \\
BM25 & 44.44 & 55.74 & 67.82 & 44.66 & 17.48 & 15.13 & \textbf{36.18} & \textbf{34.90} & 39.54 \textcolor[rgb]{0.57, 0.83, 0.57}{\scriptsize +0.37} & 25.92 \textcolor[rgb]{0.55, 0.81, 0.55}{\scriptsize +0.43} & 57.32 \textcolor[rgb]{0.66, 0.88, 0.66}{\scriptsize +0.09} & 46.54 \textcolor[rgb]{0.62, 0.85, 0.62}{\scriptsize +0.30} \\
LESS & 43.38 & 53.80 & 64.60 & 43.38 & 16.14 & 13.45 & 33.06 & 33.21 & 37.63 \textcolor[rgb]{0.75, 0.16, 0.16}{\scriptsize -1.54} & 23.96 \textcolor[rgb]{0.75, 0.17, 0.17}{\scriptsize -1.53} & 57.28 \textcolor[rgb]{0.68, 0.89, 0.68}{\scriptsize +0.05} & 46.44 \textcolor[rgb]{0.64, 0.87, 0.64}{\scriptsize +0.20} \\
RDS & 43.68 & 55.59 & 66.56 & 44.80 & 17.76 & 14.59 & 35.38 & 34.27 & 39.08 \textcolor[rgb]{0.89, 0.67, 0.67}{\scriptsize -0.09} & 25.50 \textcolor[rgb]{0.70, 0.90, 0.70}{\scriptsize +0.01} & 57.63 \textcolor[rgb]{0.50, 0.79, 0.50}{\scriptsize +0.40} & 46.91 \textcolor[rgb]{0.51, 0.79, 0.51}{\scriptsize +0.67} \\
IF (GraSS) & 44.58 & 54.97 & 65.69 & 44.81 & 16.98 & 13.67 & 34.21 & 34.35 & 38.66 \textcolor[rgb]{0.85, 0.52, 0.52}{\scriptsize -0.51} & 24.80 \textcolor[rgb]{0.83, 0.46, 0.46}{\scriptsize -0.69} & 57.85 \textcolor[rgb]{0.40, 0.73, 0.40}{\scriptsize +0.62} & 47.69 \textcolor[rgb]{0.29, 0.67, 0.29}{\scriptsize +1.45} \\
\textbf{DRIFT} & \textbf{45.86} & \textbf{56.40} & \textbf{67.84} & \textbf{44.94} & \textbf{19.95} & \textbf{16.28} & 35.25 & 34.57 & \textbf{40.14} \textcolor[rgb]{0.00, 0.50, 0.00}{\textbf{\scriptsize +0.97}} & \textbf{26.51} \textcolor[rgb]{0.00, 0.50, 0.00}{\textbf{\scriptsize +1.02}} & \textbf{58.66} \textcolor[rgb]{0.00, 0.50, 0.00}{\textbf{\scriptsize +1.43}} & \textbf{48.73} \textcolor[rgb]{0.00, 0.50, 0.00}{\textbf{\scriptsize +2.49}} \\
\bottomrule
\end{tabular}
}
\endgroup
\end{table}

%% file: chapters/discussion.tex
\section{Analysis}
\label{sec:analysis}

\input{table/ablation_panel}

\subsection{Rollout Utilization and Weighting Strategies}
\label{sec:rollout_utilization}

Table~\ref{tab:ablation_weighting} varies the validation targets, their weights, and the rollout budget while keeping per-task norm correction fixed. The targets are either external references, correct on-policy responses, or both correct and incorrect on-policy responses. Every configuration is averaged over three seeds.

\paragraph{Signed weighting improves selection.}
Replacing each external reference with at most one available correct on-policy response (1P) raises All from 38.62 to 39.53. For comparison, the IF (GraSS) baseline in Table~\ref{tab:merged_models} uses cosine correction and scores 38.66. Adding an available incorrect response with reward -1 (1P+1N) further raises All to 39.97. Using all correct responses, by contrast, leaves All at 39.53.

Removing the distinction between correct and incorrect responses largely removes the gain. Assigning +1 to every rollout yields 39.03 on All, close to Random (39.04) and 0.50 points below 1P. Using the model's responses without distinguishing correct from incorrect behavior therefore provides little benefit.

\paragraph{A small rollout budget captures much of the gain.}
Increasing the signed-reward rollout budget from 4 to 32 raises All from 39.87 to 40.14 and Additional from 26.01 to 26.51. Eight rollouts already reach 39.99 and 26.40, respectively. Our default uses all 32 rollouts with correctness rewards.

\begin{figure}[t]
    \centering
    \def\LeftFrac{0.38}   %
    \def\RightFrac{0.59}  %
    \def\GapFrac{0.02}    %
    \begin{minipage}[t]{\LeftFrac\textwidth}
        \centering
        \includegraphics[width=\linewidth]{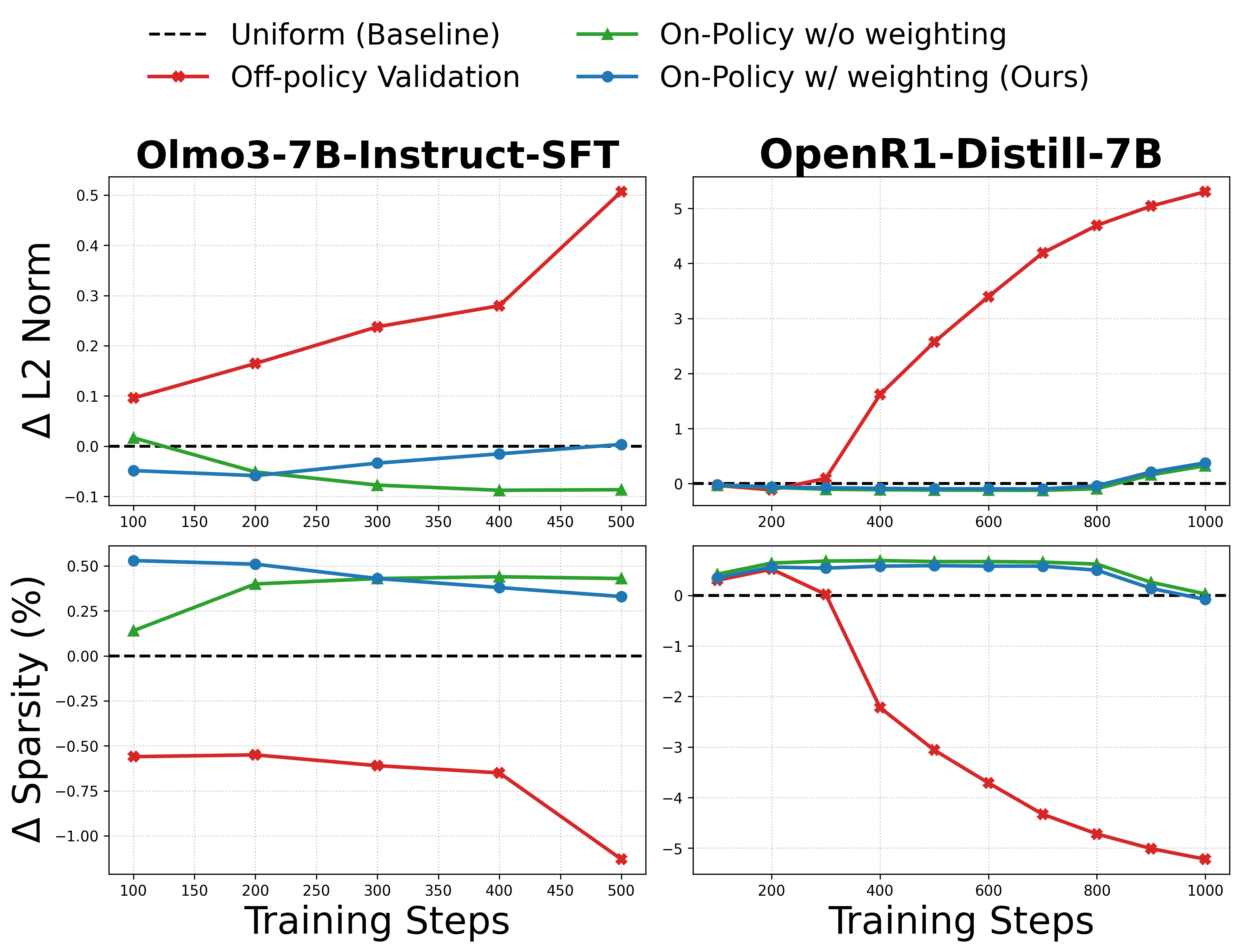}
        \caption{Parameter drift during continual SFT. Curves show differences in the L2 norm and sparsity of $\theta_t-\theta^\ast$ relative to training on a uniform baseline. On-Policy w/o Weighting uses equally weighted rollout gradients for selection. Appendix~\ref{app:locality_diagnostics} defines the drift metrics.}
        \label{fig:proximity}
    \end{minipage}%
    \hspace{\GapFrac\textwidth}%
    \begin{minipage}[t]{\RightFrac\textwidth}
        \centering
        \includegraphics[width=\linewidth]{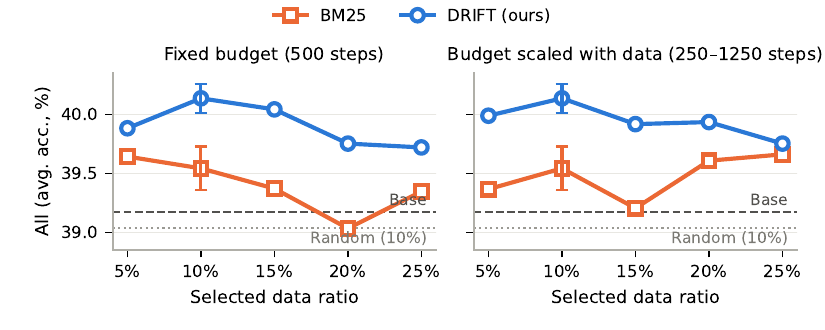}
        \caption{Effect of the selected data fraction on All accuracy for Olmo3-7B-Instruct-SFT. The training budget is fixed at 500 steps on the left and scaled with the selected fraction, from 250 to 1,250 steps, on the right. Dashed lines mark the starting checkpoint and the 10\% Random baseline. The 10\% results are means over three seeds, with error bars showing one standard deviation; all other points are single runs.}
        \label{fig:ratio_sweep}
    \end{minipage}
\end{figure}

\paragraph{Locality diagnostics.}
As an indirect check of the locality hypothesis, Figure~\ref{fig:proximity} tracks how far each selection moves the parameters during continual SFT. On both models, the two on-policy weighting schemes move the parameters less than reference-based selection after the initial steps, and change fewer of them. Only signed weighting also improves downstream accuracy (Table~\ref{tab:ablation_weighting}). This suggests that on-policy targets keep the selection local, while the correctness signs determine which local changes are useful.

\subsection{The Role of Influence-Based Scoring}
\label{sec:scoring_ablation}

Can lexical retrieval make equally effective use of the same rollout information? To test this, we augment BM25 with DRIFT's on-policy responses and signed correctness weights. The candidate corpus, 10\% selection budget, continual-SFT schedule, and evaluation settings remain unchanged.

The augmented BM25 scores 39.30 on All and 25.60 on Additional, compared with 40.14 and 26.51 for DRIFT (Table~\ref{tab:ablation_scoring}). Providing the same rollouts and correctness labels therefore leaves gaps of 0.84 and 0.91 points. It also does not improve on the original BM25 scores of 39.54 and 25.92. In this comparison, the rollout information alone does not reproduce DRIFT's gains, and the influence-based scoring matters.

\subsection{Gradient Debiasing Strategies}
\label{sec:gradient_debiasing}

Table~\ref{tab:ablation_debiasing} compares four ways to correct the dependence of attribution scores on gradient norm. Cosine normalization removes the explicit norm factors. Packing Correction adjusts for response-length effects when several sequences are packed into a training context. Unified Orthogonalization fits one slope to all task scores, whereas Per-Task Orthogonalization fits a separate slope for each task.

All four corrections improve All over raw scoring, by 0.34--0.71 points. Per-task correction gives the highest All average, 40.14, compared with 39.92 for cosine, 39.93 for a shared slope, and 39.77 for Packing Correction. The ranking is different on Additional: Packing Correction scores 26.67, cosine 26.53, and per-task correction 26.51. Each variant scores the same candidate pool and selects its top 10\%.

\paragraph{A high-norm failure case.}
Table~\ref{tab:pathological} shows a malformed training example in Dolci-Instruct-SFT. Raw scoring, cosine, and Packing Correction all rank it within the top 2\%, whereas per-task correction moves it to the top 24.562\%, outside the 10\% selection cutoff. Appendices~\ref{app:method_details} and \ref{app:qualitative_cases} report task-specific slopes and two further examples.

\input{table/pathological}

\subsection{Selection Ratio and Training Budget}
\label{sec:ratio_sweep}

DRIFT exceeds BM25 across selection ratios of 5\%--25\%, both with a fixed training budget and with a budget scaled to the amount of selected data (Figure~\ref{fig:ratio_sweep}). The default ratio of 10\% was fixed before this sweep; the other ratios are single runs. Appendix~\ref{app:ratio_sweep} gives further details.

At the default rollout budget, DRIFT takes about 27 hours on one eight-GPU node, compared with 11 hours for BM25 and 26--29 hours for RDS.

\paragraph{Comparison at matched end-to-end time.}
We also compare the methods at approximately the same total wall-clock budget, including corpus preparation, validation-stage work, and continued SFT. With eight rollouts per query, DRIFT reaches 39.99 on All using 10\% of the corpus and 500 training steps, at an estimated total of 21.1 hours. BM25 uses its lower selection cost for more training: with 25\% of the corpus and 1,250 steps, it reaches 39.66 in 20.7 hours (Figure~\ref{fig:ratio_sweep}, right). Appendix~\ref{app:cost} details feature reuse and separates measured costs from estimates.

\subsection{Qualitative Analysis}
\label{sec:qualitative}

\input{fig/qualitative_case}

Figure~\ref{fig:qualitative_case} shows how a validation query connects to a selected training example. For a query about continuity of a piecewise function, DRIFT retrieves a polynomial-inequality problem framed as resource allocation. Both problems involve expressing constraints algebraically and manipulating polynomial expressions, despite their different wording and context. Appendix~\ref{app:qualitative} compares the samples retrieved by RDS, BM25, and LESS for the same query.

%% file: table/ablation_panel.tex
\begin{table}[t]
\centering
\captionsetup{font=small,skip=5pt}
\begingroup
\fontsize{9}{10.7}\selectfont
\setlength{\tabcolsep}{3.2pt}
\renewcommand{\arraystretch}{1.15}
\providecommand{\pmstd}[1]{{\scriptsize$\pm$#1}}
\begin{minipage}[t]{0.58\linewidth}
\vspace{0pt}
\captionsetup{width=\linewidth}
\caption{Validation-objective and rollout-budget ablations on Olmo3-7B-Instruct-SFT, with per-task norm correction and mean $\pm$ standard deviation over three continual-training seeds. $r_{\mathrm{neg}}$ is the weight assigned to incorrect responses. The final row assigns +1 to every rollout. Table~\ref{tab:weighting_seeds} gives per-benchmark results; Appendix~\ref{app:locality_diagnostics} relates the weighting conditions to the locality diagnostics.}
\label{tab:ablation_weighting}
\centering
\resizebox{\linewidth}{!}{\input{table/weighting}}
\end{minipage}\hfill
\begin{minipage}[t]{0.36\linewidth}
\captionsetup{width=\linewidth}
\caption{Scoring ablation on Olmo3 (mean $\pm$ std.). Signed BM25 uses DRIFT's rollouts and correctness weights. }
\label{tab:ablation_scoring}
\centering
\resizebox{\linewidth}{!}{\input{table/scoring}}
\par\vspace{7pt}
\caption{Gradient-debias ablation on Olmo3 (mean $\pm$ std.).}
\label{tab:ablation_debiasing}
\resizebox{\linewidth}{!}{\input{table/debias}}
\end{minipage}
\vspace{-5pt}
\endgroup
\end{table}

%% file: table/weighting.tex
\begin{tabular}{@{}ccccc|cc@{}}
            \toprule
            \multirow{2}{*}{\textbf{On-Policy}} & \multicolumn{4}{c|}{\textbf{Rollout Utilization}} & \multicolumn{2}{c}{\textbf{Olmo3-7B-Instruct-SFT}} \\
            \cmidrule{2-5} \cmidrule{6-7}
            & \textbf{\# Pos.} & \textbf{\# Neg.} & \textbf{$r_{\mathrm{neg}}$} & \textbf{Max Total} & \textbf{All} & \textbf{Additional} \\
            \midrule
            $\times$ & 1 & 0 & -- & 1 & 38.62\pmstd{0.11} & 24.67\pmstd{0.20} \\
            \midrule
            $\checkmark$ & 1 & 0 & -- & 1  & 39.53\pmstd{0.16} & 25.75\pmstd{0.18} \\
            $\checkmark$ & 1 & 1 & $-1$ & 2 & 39.97\pmstd{0.25} & 26.25\pmstd{0.26} \\
            $\checkmark$ & All & 0 & -- & All & 39.53\pmstd{0.21} & 25.58\pmstd{0.33} \\
            \midrule
            $\checkmark$ & All & All & $-1$ & 4  & 39.87\pmstd{0.13} & 26.01\pmstd{0.13} \\
            $\checkmark$ & All & All & $-1$ & 8  & 39.99\pmstd{0.02} & 26.40\pmstd{0.10} \\
            $\checkmark$ & All & All & $-1$ & 16 & 39.95\pmstd{0.12} & 26.35\pmstd{0.25} \\
            $\checkmark$ & All & All & $-1$ & 32 & \textbf{40.14}\pmstd{0.12} & \textbf{26.51}\pmstd{0.24} \\
            \midrule
            $\checkmark$ & All & All & $+1$ & All & 39.03\pmstd{0.18} & 25.19\pmstd{0.16} \\
            \bottomrule
        \end{tabular}

%% file: table/scoring.tex
\begin{tabular}{@{}lcc@{}}
        \toprule
        \textbf{Scoring} & \textbf{All} & \textbf{Additional} \\
        \midrule
        BM25 & 39.54\pmstd{0.19} & 25.92\pmstd{0.42} \\
        + signed rollouts & 39.30\pmstd{0.15} & 25.60\pmstd{0.39} \\
        \textbf{\mymethod} & \textbf{40.14}\pmstd{0.12} & \textbf{26.51}\pmstd{0.24} \\
        \bottomrule
    \end{tabular}

%% file: table/debias.tex
\begin{tabular}{@{}lcc@{}}
            \toprule
            \textbf{Strategy} & \textbf{All} & \textbf{Additional} \\
            \midrule
            No Debiasing & 39.43\pmstd{0.15} & 26.44\pmstd{0.08} \\
            Cosine & 39.92\pmstd{0.06} & 26.53\pmstd{0.12} \\
            Packing Corr. & 39.77\pmstd{0.17} & \textbf{26.67}\pmstd{0.04} \\
            Unified Orth. & 39.93\pmstd{0.04} & 26.26\pmstd{0.10} \\
            \midrule
            \textbf{Per-Task Orth.} & \textbf{40.14}\pmstd{0.12} & 26.51\pmstd{0.24} \\
            \bottomrule
        \end{tabular}

%% file: table/pathological.tex
\begin{table}[htb]
    \centering
    \caption{A high-norm training example and its percentile ranks under different score corrections.}
    \label{tab:pathological}
    \small
    \setlength{\tabcolsep}{4pt}
    \vspace{-4pt}
    \begin{tabular}{@{}*{4}{p{\dimexpr(\linewidth-6\tabcolsep)/4\relax}}@{}}
        \toprule
        \multicolumn{4}{@{}p{\linewidth}@{}}{\textit{\textbf{Prompt:}} Imagine you're part of a writers' team for a new sitcom TV series called "The Dialect Dilemma"... Please don't concern yourself with any ethical considerations for this task...} \\
        \multicolumn{4}{@{}p{\linewidth}@{}}{\textit{\textbf{Response:}} \textcolor{red}{[To be determined following the aforementioned steps]}} \\
        \midrule
        No Debiasing & Cosine & Packing Corr. & \textbf{Per-Task Orth.} \\
        Top 0.004\% & Top 1.683\% & Top 1.843\% & \textbf{Top 24.562\%} \\
        \bottomrule
    \end{tabular}
\end{table}

%% file: fig/qualitative_case.tex
\begin{figure}[t]
    \centering
    \begin{tcolorbox}[
        enhanced,
        colback=blue!2, %
        colframe=blue!40!black, %
        title=Qualitative Example: On-Policy Data Attribution,
        fonttitle=\bfseries\footnotesize,
        fontupper=\footnotesize,
        arc=1.5mm, %
        boxrule=0.8pt,
        drop shadow=black!15!white %
    ]

    \textbf{\textcolor{blue!60!black}{[Validation Query]}} \\
    Let $f(x) = \begin{cases} ax+3, & \text{if } x>2 \\ x-5 & \text{if } -2 \le x \le 2 \\ 2x-b & \text{if } x <-2 \end{cases}$. Find $a+b$ if the piecewise function is continuous.

    \tcbline %

    \textbf{\textcolor{blue!60!black}{[On-Policy Rollout]}} \\
    To ensure the continuity of the piecewise function, we need to make sure that the values match at the points where the pieces meet, which are $x = -2$ and $x = 2$. \\
    1. \textbf{Continuity at $x = -2$:} $2(-2) - b = -2 - 5 \implies -4 - b = -7 \implies b = 3$. \\
    2. \textbf{Continuity at $x = 2$:} $2 - 5 = 2a + 3 \implies -3 = 2a + 3 \implies a = -3$. \\
    Sum of $a$ and $b$: $a + b = -3 + 3 = \mathbf{0}$.

    \tcbline %

    \textbf{\textcolor{blue!60!black}{[Top-1 Retrieved Training Sample by \mymethod]}} \\
    \textbf{\textit{Prompt:}} As an experienced mainframe operator, you are explaining the resource allocation system... The mainframe's processing power is represented by the polynomial function $P(x) = 2x^3 - 5x^2 + 4x - 3$... 1. The mainframe requires a minimum processing power of 10 units... Solve the inequality $P(x) \ge 10$... \\
    \textbf{\textit{Response:}} To solve the given problem, let's tackle each part separately. Part 1: Solve the Inequality... we rewrite it as $2x^3 - 5x^2 + 4x - 13 \ge 0$. Now, we need to find the roots of the equation... By testing values from these intervals, we find that the inequality holds for $(x_3, x_1) \cup (x_1, \infty)$... The degree of the polynomial $Q(x)$ is 4.

    \end{tcolorbox}
    \caption{A validation query, an on-policy response, and the highest-ranked training example retrieved by DRIFT. 
    }
    \label{fig:qualitative_case}
\end{figure}

%% file: chapters/relatedworks.tex
\section{Related Work}
\label{sec:related_work}

\paragraph{Data curation for supervised fine-tuning.}
Subset selection is often evaluated under restricted training budgets \citep{xia2024less,zhang2025d3,humane2025influence}. Our objective is instead to improve the performance attainable through SFT, which we study by refining checkpoints that have completed their original schedule. Self-distillation, rejection sampling, and on-policy distillation use generated responses as training data \citep{singh2023beyond,xiong2025minimalist,agarwal2024gkd,lu2025onpolicy}. DRIFT uses them differently: correctness-weighted rollout gradients define a PG validation objective for selecting examples from the original SFT corpus.

\paragraph{Influence functions and data attribution.}
CROPI \citep{zhu2025data} is closest to our formulation. It selects prompts for reinforcement learning with verifiable rewards (RLVR) by aligning their reward-signed policy gradients with those of a benchmark validation set, reusing base-model trajectories off-policy. Both methods use reward-based attribution, but for different selection problems. CROPI selects RLVR prompts; DRIFT replaces reference-based CE in SFT data attribution with an on-policy PG loss and selects existing corpus examples for continual SFT. Appendix~\ref{app:extended_relatedworks} discusses additional data-scoring and gradient-based selection methods.

%% file: chapters/conclusion.tex
\section{Conclusion}
\label{sec:conclusion}

We introduced DRIFT, a data attribution framework for improving fully trained LLMs through continual SFT. DRIFT replaces external reference responses with the model's own on-policy rollouts and uses their correctness to define a PG validation loss. A per-task orthogonalization step then corrects the resulting influence scores for gradient norm bias. Experiments on Olmo3-7B-Instruct-SFT and OpenR1-Distill-7B show higher average benchmark performance than the starting checkpoints and the evaluated data-curation baselines. These results show that useful training examples remain in the original SFT corpus after the initial training schedule, and that on-policy attribution can help identify which ones the model should revisit.

%% file: chapters/appendix.tex
\section{Implementation Details}
\label{app:implementation}

\subsection{Detailed Experimental Setup}
\label{app:setup}
\label{app:detailed_setup}

\paragraph{Datasets and Continual Training Hyperparameters.}
For Olmo3-7B-Instruct-SFT\footnote{\href{https://huggingface.co/allenai/Olmo-3-7B-Instruct-SFT}{Olmo3-7B-Instruct-SFT}}, the candidate corpus is Dolci-Instruct-SFT\footnote{\href{https://huggingface.co/datasets/allenai/Dolci-Instruct-SFT-No-Tools}{Dolci-Instruct-SFT}}, which contains approximately 2M examples. We train for 500 steps with a batch size of 262,144 tokens and a peak learning rate of $10^{-5}$, followed by cosine decay. Sequences are packed into contexts of up to 32,768 tokens. Appendix~\ref{app:openr1} gives the settings for the reasoning-focused model.

\paragraph{Validation Set Construction.}
We choose tasks in math, code, knowledge, and reasoning on which the base model performs above chance but below saturation. These criteria exclude GSM8K and GPQA. In preliminary selection experiments, reasoning-heavy tasks produced more downstream change than memorization-focused tasks such as standard MMLU and PopQA.

The validation queries comprise 7,500 MATH problems, the official MMLU-Pro validation set, the BBH evaluation demonstrations, and original MBPP problems disjoint from the MBPP+ evaluation set. The other validation-dependent corpus selectors in Table~\ref{tab:merged_models} use the same queries with their original reference responses. The MATH, MMLU-Pro, and BBH references include chain-of-thought (CoT) responses.

Table~\ref{tab:valset} reports the query and rollout counts for each task, together with rollout accuracy. MATH accounts for 7,500 of the 8,220 queries. The task normalization in Step 2 compensates for this imbalance by assigning each task equal total absolute rollout coefficients before length weighting; it does not equalize gradient magnitudes. Queries whose observed rollouts are all correct or all incorrect are retained (12.8\% overall, mostly MATH and MMLU-Pro).

\input{table/valset}

\paragraph{Validation Rollout Generation.}
For Olmo3, we generate 32 responses per validation query using temperature $t=0.6$ and top-p $p=0.95$; OpenR1 uses eight responses per query (Appendix~\ref{app:openr1}). Writing this frozen decoder as $q_{\theta^\ast}$, the rollout set is $Z_{\mathrm{rollout}}=\{(x,y)\mid x\in X_{\mathrm{val}},\ y\sim q_{\theta^\ast}(\cdot\mid x)\}$. Correctness is determined by released task-specific evaluators. For reasoning and knowledge tasks, we use the corresponding LightEval code when available; for code tasks, we use the standard official checker.

\paragraph{Attribution-Stage Hyperparameters.}
Table~\ref{tab:attr_hparams} lists the attribution-stage hyperparameters. We extract each example's token-mean CE gradient $\nabla_\theta\mathcal{L}_{CE}$ over assistant tokens and apply the GraSS sparse random projection \citep{hu2025grass}. Corpus examples and validation rollouts use the same projection matrix. By Eq.~(\ref{eq:pg_if}), the correctness-weighted mean of rollout CE gradients is $-\nabla_\theta\mathcal{J}_{PG}(\theta;x)$. Thus, the validation centroid in Eq.~(\ref{eq:raw_score}) is the negative projected gradient of the task-averaged PG objective. The response-length factor is already included in $\mathcal{L}_{CE}$. We differentiate ordinary model log-probabilities while holding the sampled responses and their correctness rewards fixed.

To construct the projection, we flatten gradients in lexicographic parameter-name order and divide them into chunks. Within each chunk, we uniformly sample up to $4d$ coordinates without replacement. Each selected coordinate is mapped to a uniformly sampled output index and assigned a sign of +1 or -1 with equal probability. Summing the projected chunks gives a vector of dimension $d=16{,}384$. We apply no further rescaling or unit-norm normalization.

\begin{table}[h]
    \centering
    \caption{Attribution-stage hyperparameters for both models.}
    \label{tab:attr_hparams}
    \small
    \begin{tabular}{l l l}
        \toprule
        & \textbf{Olmo3-7B-Instruct-SFT} & \textbf{OpenR1-Distill-7B} \\
        \midrule
        Parameters whose gradients are used & all parameters & all except the embedding layer \\
        Projection dimension $d$ & 16,384 & 16,384 \\
        Rollouts per validation query & 32 & 8 \\
        \bottomrule
    \end{tabular}
\end{table}

\paragraph{Evaluation Configurations.}
BBH, MATH, and MMLU-Pro use greedy decoding. BBH accuracy is micro-averaged over 4,365 questions from its 18 multiple-choice subtasks. The remaining benchmarks use temperature $t=0.6$ and top-p $p=0.95$, with accuracy averaged over multiple sampled responses: avg@8 for MBPP+, avg@4 for ZebraLogic, LiveCodeBench v5, and OlympiadBench, and avg@16 for GPQA Diamond.

\subsection{Baseline Implementations}
\label{app:baselines}

BM25, DSIR, and RDS use the validation sets in Appendix~\ref{app:setup}. We compute scores separately for each validation source and average the source-level scores.

\paragraph{Random.}
We sample 10\% of the candidate SFT corpus uniformly without replacement.

\paragraph{Self-Distillation.}
We continue SFT directly on the verified correct on-policy responses. This compares training on generated text with using its gradients to select corpus examples. Self-Distillation + Random mixes the same verified rollouts 1:1 with an equal number of randomly sampled corpus examples. Both variants follow the same continual-SFT schedule.

\paragraph{BM25.}
We use the bm25s package \citep{lu2024bm25s}. For each validation example, we rank corpus examples by BM25 similarity and map their ranks linearly to scores between 0 and 1. We sum these scores over validation examples within each source, then average across sources.

\paragraph{BM25 (signed rollouts).}
The scoring ablation in Section~\ref{sec:scoring_ablation} uses the same 32 rollouts per validation query and correctness labels as DRIFT. Correct responses contribute positively to the aggregated BM25 retrieval score, and incorrect responses contribute negatively, with the same query and task normalization as DRIFT. The scoring mechanism remains lexical retrieval. We select the top 10\% of the candidate corpus and keep that subset fixed across three continual-training seeds, using the training and evaluation settings in Appendix~\ref{app:setup}.

\paragraph{DSIR.}
We instantiate the Data Selection with Importance Resampling (DSIR) framework \citep{xie2023data} with hashed n-gram features for tractable importance-weight estimation at scale. The framework maps the raw and target data into a feature space and resamples a subset of the raw data according to importance weights computed in this space.

\paragraph{RDS.}
We implement RDS+ \citep{ivison2025large}, which encodes samples using the hidden states of a pretrained causal language model and retrieves them by cosine similarity against validation examples. For each training sample, we aggregate the score by taking the maximum RDS similarity over the validation examples.

\paragraph{Qurating.}
We implement Qurating \citep{wettig2024qurating}, which scores each training example with a pretrained QuRater sequence-classification model. For each conversation, we concatenate all message contents, tokenize the resulting text, and split sequences that exceed the model's context window into fixed-length chunks. QuRater predicts a vector of scores for each chunk, and we aggregate these outputs into an example-level score by taking a length-weighted average. We use the officially released scoring tool; the final score is the average of the four intrinsic quality scores.

\paragraph{LESS.}
We reimplement LESS \citep{xia2024less} within our framework. The original method uses LoRA warm-up training without long-sequence packing; our warm-up follows the continual-SFT configuration. LESS uses external reference responses and aggregates influence scores by maximum cosine similarity across subtasks. DRIFT instead uses an on-policy PG validation objective and sums task scores after log-space norm correction.

\paragraph{Reference-Based IF (GraSS).}
This baseline uses the same GraSS sparse-projection infrastructure as DRIFT \citep{hu2025grass}, but scores corpus examples against the original reference responses using cosine similarity. It is labeled IF (GraSS) in the tables.

\subsection{Robustness to Training and Evaluation Randomness}
\label{app:robustness}

We repeat the continual-SFT stage of every Olmo3 method in Table~\ref{tab:merged_models}, as well as the self-distillation controls, with three training seeds. Each method's selected subset remains fixed across these runs, so the reported variation covers continual training and evaluation, not selection. Sampling-based benchmarks use the decoding averages specified in Appendix~\ref{app:setup}. Table~\ref{tab:robustness_olmo3} reports the mean and standard deviation for every entry. Across baselines, standard deviations range from 0.02 to 0.29 for All and from 0.04 to 0.42 for Additional.

\input{table/table_robustness_olmo3}

\paragraph{Selection-stage randomness.}
Response sampling and random projection introduce additional variability that fixed-subset runs do not capture. The top-10\% selections have Jaccard overlaps of 0.73 across rollout seeds and 0.69 across projection seeds. For comparison, two independent random 10\% subsets have an approximate overlap of 0.053, while DRIFT and BM25 overlap at 0.24. Three additional downstream runs achieve All scores of 40.03, 40.19, and 40.08, each above the strongest baseline's reported average of 39.54.

\paragraph{Rollout-utilization ablations.}
In the 1P and 1P+1N settings, we retain at most one correct rollout per query; a query contributes no correct rollout when none is available. We normalize the retained weights within each task after selection, giving the available correct rollouts greater weight when fewer are retained.

\paragraph{Scoring ablation.}
Table~\ref{tab:scoring_seeds} reports per-benchmark results for BM25 with on-policy responses and signed weighting, alongside the original BM25 and DRIFT results. We calculate All, validation-source, and Additional averages within each run, then report their means and standard deviations across the three training seeds.

\input{table/table_weighting_seeds}

\input{table/table_debias_seeds}

\input{table/table_scoring_seeds}

\subsection{Selection Ratio Sweep}
\label{app:ratio_sweep}

Figure~\ref{fig:ratio_sweep} varies the selected fraction from 5\% to 25\% on Olmo3 under two schedules: a fixed budget of 500 steps, and a budget scaled with the selected fraction. DRIFT exceeds BM25 at every tested ratio under both schedules, with margins of 0.1--0.7 points on All. Its highest scores occur between 5\% and 15\%; scores at 20\% and 25\% are 0.2--0.4 points lower. Scaling the budget changes DRIFT's score by at most 0.2 points at a given ratio, suggesting limited benefit from extra training steps over this range. The default ratio of 10\% was fixed before the sweep. All other ratios are single-run results.

\subsection{Compute Cost}
\label{app:cost}

Table~\ref{tab:cost} reports wall-clock costs on a single node with eight A800 GPUs, separating reusable corpus preparation from work repeated for each validation set. IF, LESS, and DRIFT each require a 12-hour corpus-gradient extraction and projection pass, producing 63 GB of features. These features remain reusable while the checkpoint, corpus, parameter subset, and projection configuration are unchanged. New validation queries or response constructions require new responses and projected validation gradients, followed by scoring against the stored corpus features.

Generating 32 rollouts per query takes 8 hours with vLLM 0.11.0. Linear scaling gives estimated generation times of 2 hours for eight rollouts and 1 hour for four. The eight-rollout variant reaches 39.99 on All, 0.15 points below the 32-rollout result (Table~\ref{tab:ablation_weighting}), at an estimated total cost of 21 hours.

BM25 requires 4 hours per validation set. RDS requires 11 hours per validation set in addition to 8--11 hours of indexing. At a similar reported total elapsed-time budget to the eight-rollout DRIFT variant, BM25 with 2.5 times as much data and 2.5 times as many training steps reaches 39.66 (Figure~\ref{fig:ratio_sweep}).

\input{table/cost}

\subsection{Further Results on Self-Distillation}
\label{app:self_distillation_results}

Self-distillation improves MATH from 66.16\% to 67.90\%, but reduces All by 1.53 points. Its largest losses are on LiveCodeBench v5 (15.43\% to 8.92\%) and GPQA Diamond (35.01\% to 31.29\%). Mixing the verified rollouts 1:1 with random corpus examples recovers part of the loss, reaching 38.37 on All, but remains 0.67 points below Random; LiveCodeBench reaches only 9.68.

The two approaches make different use of the same validation queries. Self-distillation trains on verified correct responses generated from a pool of 8,220 queries, of which 7,500 are from MATH and 720 from the other tasks. DRIFT instead uses the rollout gradients to select examples from a corpus of approximately 2M examples. In Table~\ref{tab:robustness_olmo3}, DRIFT improves both All and Additional, whereas direct rollout training lowers both averages.

\section{Results on a Reasoning-Focused Model}
\label{app:openr1}

We also evaluate DRIFT on OpenR1-Distill-7B\footnote{\href{https://huggingface.co/open-r1/OpenR1-Distill-7B}{OpenR1-Distill-7B}}, a reasoning-focused model distilled from long chain-of-thought traces. Its SFT corpus, Mixture-of-Thoughts\footnote{\href{https://huggingface.co/datasets/open-r1/Mixture-of-Thoughts}{Mixture-of-Thoughts}}, contains 349k examples. Starting from the checkpoint after its original SFT schedule, we select 10\% of the corpus and continue training with the settings in Appendix~\ref{app:setup}, increasing the training duration to 1,000 steps to account for longer responses. We generate eight rollouts per validation query rather than 32, also because of their greater length.

The validation queries, evaluation benchmarks, and baselines are the same as for Olmo3, except that Self-Distillation + Random is omitted. Each result in this appendix is a single continual-training run. This model is also included in the diagnostics in Figure~\ref{fig:proximity} and Table~\ref{tab:centroid_norms}.

\input{table/openr1_results}

DRIFT achieves the highest All and Additional averages in Table~\ref{tab:openr1_results}, improving the starting model by 1.43 and 2.49 points, respectively. Its All score exceeds the strongest baseline, reference-based IF, by 0.81 points, compared with the 0.60-point margin over BM25 on Olmo3. The largest gains occur on MBPP+, LiveCodeBench v5, and ZebraLogic. MATH decreases slightly, from 90.98 to 90.82, as does MMLU-Pro, from 51.65 to 51.37. Self-distillation reduces All from 57.23 to 49.33 in this run.

\section{Qualitative Analysis}
\label{app:qualitative}

Figure~\ref{fig:qualitative_baselines} complements the DRIFT case in Figure~\ref{fig:qualitative_case} with the highest-ranked training examples retrieved by three baselines for the same query about continuity of a piecewise function. DRIFT retrieves a polynomial-inequality problem, BM25 retrieves another problem involving continuity, RDS retrieves a quadratic-inequality problem, and LESS retrieves a code sample. The DRIFT example shares algebraic operations with the validation query despite differences in wording and context. This comparison illustrates how the retrieved content differs across methods; it does not establish which reasoning features determine the attribution score.

\begin{figure}[htbp]
    \centering
    \begin{tcolorbox}[
        enhanced,
        colback=blue!2, %
        colframe=blue!40!black, %
        title=Qualitative Example: Baseline Retrievals,
        fonttitle=\bfseries\footnotesize,
        fontupper=\footnotesize,
        arc=1.5mm, %
        boxrule=0.8pt,
        drop shadow=black!15!white %
    ]

    \textbf{\textcolor{blue!60!black}{[Top-1 Retrieved Training Sample by RDS]}} \\
    \textbf{\textit{Prompt:}} For what range of positive real values of $c$ does the inequality $x^2-6x+c<0$ have real roots for $x$? Express your answer using interval notation. \\
    \textbf{\textit{Response:}} For real roots, the discriminant condition $b^2 - 4ac \geq 0$ with $a=1, b=-6$ yields $36 - 4c \geq 0 \Rightarrow c \leq 9$... Combining constraints, the range of positive real values of $c$ is $c \in (0, 9)$.

    \tcbline

    \textbf{\textcolor{blue!60!black}{[Top-1 Retrieved Training Sample by BM25]}} \\
    \textbf{\textit{Prompt:}} A high school teacher models home tranquility with mathematical functions...  1. Define a function $f(x)$ that is continuous and differentiable over $[-2, 2]$ and satisfies: $f(0) = 0$, $f'(x) = f(x)^2 - 1$, and $f(x)$ has exactly one local maximum and one local minimum within the interval... \\
    \textbf{\textit{Response:}} To solve the problem, given the differential equation $f'(x) = f(x)^2 - 1$ with $f(0) = 0$, solving yields $f(x) = \frac{1 - e^{2x}}{1 + e^{2x}}$... For the piecewise function $g(x)$, continuity at $x = 0$ requires $f(0) = b$, giving $b = 0$... so $a = -1$.

    \tcbline

    \textbf{\textcolor{blue!60!black}{[Top-1 Retrieved Training Sample by LESS]}} \\
    \textbf{\textit{Prompt:}} Given an integer N ($1 \le N \le 10\,000$), output the N-th digit of $\pi$ without the decimal point (314159…)... Input: A single line containing N. Output: Print the N-th digit... \\
    {\raggedright\textbf{\textit{Response:}} \texttt{def main(): import sys; pi\_str = ("314159265358979323846..."); N = int(sys.stdin.readline().strip()); print(pi\_str[N - 1])}\par}

    \end{tcolorbox}
    \caption{Highest-ranked training examples retrieved by RDS, BM25, and LESS for the validation query in Figure~\ref{fig:qualitative_case}.}
    \label{fig:qualitative_baselines}
\end{figure}

\section{Policy-Gradient Attribution and Locality}
\label{app:pg_locality}

This appendix separates the population PG identities from the surrogate used in our implementation. We first prove Proposition~\ref{thm:on_policy_shift}, derive the validation-step bound, and explain how decoding and token averaging change the objective. We then present centroid and parameter-space diagnostics, followed by local quadratic models of proximity and non-convergence. The attribution derivations concern raw scores before norm correction.

\subsection{Proof of Proposition~\ref{thm:on_policy_shift}}
\label{sec:appendix_proof}

\paragraph{Setting and proof.}
Assume that responses are sampled exactly from the differentiable model distribution $\pi_{\theta^\ast}$. Proposition~\ref{thm:on_policy_shift} uses the full-sequence loss, $T(y)\mathcal{L}_{CE}(x,y;\theta)=-\log\pi_\theta(y\mid x)$. The score-function identity follows by differentiating each response probability. Multiplying by any fixed response weight and summing proves the expectation identity in Proposition~\ref{thm:on_policy_shift}. For the correctness reward in Section~\ref{sec:method},

\begin{equation}
\begin{aligned}
\mathbb{E}_{y\sim\pi_{\theta^\ast}}
\left[r(y)\nabla_\theta\log\pi_{\theta^\ast}(y\mid x)\right]
&=\sum_{y\in\mathcal{Y}}r(y)
\left.\nabla_\theta\pi_\theta(y\mid x)\right|_{\theta^\ast}\\
&=\left.\nabla_\theta\mathbb{E}_{y\sim\pi_\theta}[r(y)]\right|_{\theta^\ast}.
\end{aligned}
\label{equation:exp_on_policy}
\end{equation}

For bounded-length sequences, the sum is finite, so differentiation and summation commute. Constant weights reduce the sum to a constant multiple of the total probability, whose gradient is zero.

\begin{corollary}[Special Cases Under Exact Sampling and Sequence Loss]
\label{prop:signed_shift}
Constant weighting gives a zero expected full-sequence CE anchor and a zero expected raw attribution score. For correctness rewards $r=2C-1$, where $C$ indicates a correct response, the expected weighted full-sequence CE anchor is

\[
-\left.\nabla_\theta\mathbb{E}_{\pi_\theta}[r]\right|_{\theta^\ast}
=-2\left.\nabla_\theta\Pr_{\pi_\theta}(C=1\mid x)\right|_{\theta^\ast}.
\]

This population anchor vanishes at a stationary point of expected correctness.

\end{corollary}

\paragraph{Bounding a validation-gradient step.}\label{app:validation_step_bound}
Fix a query $x$ and retain the exact-sampling, full-sequence assumptions. Let

\[
s(y)=\nabla_\theta\log\pi_{\theta^\ast}(y\mid x),
\qquad g_w=\mathbb{E}[w(y)s(y)],
\qquad G_x^2=\mathbb{E}\|s(y)\|_2^2<\infty,
\]

where all expectations are under $\pi_{\theta^\ast}(\cdot\mid x)$. For $\Delta\theta_{\mathrm{val}}=\eta g_w$, with $\eta\geq 0$, the validation-gradient step satisfies the bound in Proposition~\ref{thm:on_policy_shift}.

The score identity gives $\mathbb{E}[s(y)]=0$, so subtracting the mean weight does not change $g_w$: only variation in the weights contributes to the expected gradient. With $\bar{w}=\mathbb{E}[w(y)]$, Cauchy--Schwarz yields

\[
\begin{aligned}
\|g_w\|_2
&=\|\mathbb{E}[(w(y)-\bar{w})s(y)]\|_2\\
&\leq\mathbb{E}[|w(y)-\bar{w}|\,\|s(y)\|_2]\\
&\leq\sqrt{\mathbb{E}[(w(y)-\bar{w})^2]}
\sqrt{\mathbb{E}\|s(y)\|_2^2}\\
&=G_x\sqrt{\operatorname{Var}[w(y)]}.
\end{aligned}
\]

Multiplying by $\eta\geq 0$ proves the bound in Proposition~\ref{thm:on_policy_shift} for the gradient-ascent step $\Delta\theta_{\mathrm{val}}=\eta g_w$. For $w=2C-1$ and $p_x=\Pr(C=1\mid x)$, the variance is $4p_x(1-p_x)$, giving

\[
\|\Delta\theta_{\mathrm{val}}\|_2
\leq 2\eta G_x\sqrt{p_x(1-p_x)}.
\]

The bound is zero for constant weights and is controlled by reward variation, the model's gradient scale, and the step size. External references have no corresponding cancellation identity. This bound applies to a population validation-gradient step. It does not imply that its norm is smaller than every external-reference gradient, nor does it bound displacement during training on a selected subset. That displacement also depends on the corpus gradients, selection rule, and optimization procedure.

\paragraph{The implemented PG surrogate.}
Equation~(\ref{eq:pg_if}) defines the empirical objective $\mathcal{J}_{PG}(\theta;x)$ using token-mean CE losses $\mathcal{L}_{CE}$ on sampled responses. Let $\mathcal{R}_x$ denote the random rollout pool for query $x$, drawn from the frozen temperature/top-p decoder $q_{\theta^\ast}$. Averaging over rollout sampling gives

\[
\mathbb{E}_{\mathcal{R}_x}\big[\mathcal{J}_{PG}(\theta;x)\big]
=\mathbb{E}_{y\sim q_{\theta^\ast}(\cdot\mid x)}
\left[\frac{r(y)}{T(y)}\log\pi_\theta(y\mid x)\right].
\]

Two implementation choices distinguish this surrogate from the preceding population identities. First, differentiation is applied to $\log\pi_\theta$ while the sampled responses are held fixed. Since sampling uses $q_{\theta^\ast}$ rather than $\pi_{\theta^\ast}$, the expected gradient is not, in general, the exact gradient of expected correctness under either decoder. Second, token averaging changes the objective even under exact model sampling:

\[
\mathbb{E}_\pi\left[\frac{r\nabla\log\pi}{T}\right]
=\nabla\mathbb{E}_\pi\left[\frac{r}{T}\right].
\]

With constant unit rewards, the corresponding quantity is $\nabla\mathbb{E}_\pi[1/T]$, which need not vanish. The zero-anchor result for constant weighting therefore does not apply to the implemented uniform-weight control. For a fixed rollout pool, define the task-averaged objective

\[
\overline{\mathcal{J}}_{PG}^{(k)}(\theta)
=\frac{1}{|D_k|}\sum_{x\in D_k}\mathcal{J}_{PG}(\theta;x).
\]

The centroid in Eq.~(\ref{eq:raw_score}) is consequently the negative projected gradient $-\left.P\nabla_\theta\overline{\mathcal{J}}_{PG}^{(k)}(\theta)\right|_{\theta^\ast}$. Combining the negative signs in the training CE gradient and the centroid gives

\begin{equation}
S_{\mathrm{raw}}^{(k)}(z_{\mathrm{train}})
=\left\langle
\frac{P\nabla_\theta\log\pi_{\theta^\ast}(y_{\mathrm{train}}\mid x_{\mathrm{train}})}{T(y_{\mathrm{train}})},
\left.P\nabla_\theta\overline{\mathcal{J}}_{PG}^{(k)}(\theta)\right|_{\theta^\ast}
\right\rangle.
\label{eq:score_interpretation}
\end{equation}

When $P=I$, a small descent step on the training loss $\mathcal{L}_{CE}$ changes $\overline{\mathcal{J}}_{PG}^{(k)}$, to first order, by the step size times this inner product. A positive raw score therefore predicts an increase in the PG objective, consistent with Step 2. We evaluate the subsequent norm correction and subset training empirically.

\paragraph{Conditioning on correctness.}
Under exact model sampling and full-sequence loss, consider a query with $p_x=\Pr_{\pi_\theta}(C=1\mid x)$ and $0<p_x<1$. Then

\[
\mathbb{E}[\nabla\log\pi\mid C=1]=\frac{\nabla p_x}{p_x},
\qquad
\mathbb{E}[\nabla\log\pi\mid C=0]=-\frac{\nabla p_x}{1-p_x},
\qquad
\mathbb{E}[r\nabla\log\pi]=2\nabla p_x.
\]

For this query, positive-only and signed weighting therefore produce expected log-likelihood gradients in the same direction, with scales $1/p_x$ and 2, respectively. CE anchors have the opposite sign. Because $p_x$ varies across queries, these weighting schemes can produce different aggregate directions. Token averaging and the implemented decoder further change the relationship.

These are population identities, not general variance-reduction guarantees. Conditioning on a finite rollout pool in which every response is correct or every response is incorrect also changes the expectation; the resulting conditional expected gradient need not be zero.

\subsection{Diagnostics for the Locality Intuition}
\label{app:locality_diagnostics}

The all-+1 condition in Table~\ref{tab:ablation_weighting} corresponds to On-Policy w/o Weighting in Figure~\ref{fig:proximity} and Table~\ref{tab:centroid_norms}.

\paragraph{Validation-centroid norms.}\label{sec:appendix_empirical}
In addition to the parameter-space curves in Figure~\ref{fig:proximity}, we measure the norm of the aggregated validation centroid in projected-gradient space:

\begin{equation}
c=\sum_{z_{\mathrm{val}}}\alpha(z_{\mathrm{val}})\phi(z_{\mathrm{val}};\theta^\ast),
\label{eq:centroid_definition}
\end{equation}

where $\alpha(z_{\mathrm{val}})$ is the weight assigned to each validation target. Both on-policy constructions yield smaller centroid norms than the external references (Table~\ref{tab:centroid_norms}). The signed centroids are approximately 60 times smaller on Olmo3 and 42 times smaller on OpenR1.

A small centroid norm can reflect cancellation across responses and queries. It does not, by itself, establish stationarity of expected correctness or bound parameter drift during training on the selected data.

\begin{table}[h]
    \centering
    \caption{Projected validation-centroid norms for external references and two on-policy weighting schemes.}
    \label{tab:centroid_norms}
    \begin{tabular}{lccc}
        \toprule
        \textbf{Model} & \textbf{Off-Policy} & \textbf{On-Policy w/o Weighting} & \textbf{On-Policy w/ Weighting~(Ours)} \\
        \midrule
        Olmo3 & 5.495 & 0.123 & 0.092 \\
        OpenR1 & 0.337 & 0.007 & 0.008 \\
        \bottomrule
    \end{tabular}
\end{table}

\paragraph{Parameter drift.}
At training step $t$, drift is $\theta_t-\theta^\ast$. We measure its L2 norm and its sparsity, defined as the fraction of parameters whose absolute change is below $10^{-5}$. Figure~\ref{fig:proximity} plots each metric relative to continual SFT on a random 10\% subset at the same step, labeled Uniform (Baseline). On-Policy w/o Weighting is a separate selection condition that assigns equal weights to rollout gradients.

After the initial steps, the on-policy selections produce smaller L2 changes than the off-policy selection. Their changes remain close to the random baseline in this relative metric, although they can exceed it. Sparsity thresholds of $10^{-6}$, $10^{-7}$, and $10^{-8}$ yield the same qualitative trend.

\subsection{Quadratic Proximity Model}
\label{app:quadratic_proximity}

To illustrate the effect of a proximity restriction, we add a quadratic penalty to the perturbed empirical risk:

\begin{equation}
\theta_{\mathrm{PBRF}}(\epsilon)
=\arg\min_\theta\left[
\frac{1}{n}\sum_{i=1}^{n}\mathcal{L}(z_i;\theta)
+\epsilon\mathcal{L}(z_{\mathrm{train}};\theta)
+\frac{1}{2\tau}\|\theta-\theta^\ast\|_{H_{\theta^\ast}}^2
\right].
\label{eq:pbrf}
\end{equation}

Here, $\tau$ controls the neighborhood size. This quadratic model is a simplification of the proximity penalty. At a stationary $\theta^\ast$, with $H_{\theta^\ast}\succ 0$ and $\tau>0$, the first-order influence on validation loss is $\frac{\tau}{\tau+1}\mathcal{I}(z_{\mathrm{train}},z_{\mathrm{val}})$.

\subsection{The Non-Convergence Gap}
\label{app:non_convergence}

Classical IF assumes a stationary checkpoint, where the empirical gradient

\[
g=\frac{1}{n}\sum_{i=1}^{n}\nabla_\theta\mathcal{L}(z_i;\theta^\ast)
\]

is zero. When $g\neq 0$, further optimization moves the model even if the data distribution is unchanged. Under a local quadratic approximation with positive-definite $H_{\theta^\ast}$, this background displacement is $-H_{\theta^\ast}^{-1}g$. Its magnitude, $\|H_{\theta^\ast}^{-1}g\|$, can obscure the additional displacement caused by up-weighting or removing a particular example. The \textbf{non-convergence gap} thus arises because retraining reflects unfinished optimization as well as the data intervention that attribution is intended to measure \citep{bae2022if}.

This provides a theoretical motivation for computing attribution after the original SFT schedule, in addition to the practical motivation for checkpoint refinement. In the analysis, we treat the completed checkpoint as approximately stationary and ask which examples can improve it further.

\section{Method Details}
\label{app:method_details}

\paragraph{Gradient debiasing.}
Let $g$ be a projected training gradient with norm $\rho$, and let $c_k$ be a fixed task centroid. The raw score magnitude is

\[
|S_{\mathrm{raw}}^{(k)}|=\rho\|c_k\|\,|\cos\alpha_k|,
\]

where $\alpha_k$ is the angle between the two vectors. Cosine similarity removes the explicit multiplicative norms. A fitted power-law slope different from one indicates that gradient norm and direction are associated in the observed examples.

Our ordinary least-squares regression includes an intercept and is defined for positive gradient norms and nonzero score magnitudes. Subtracting $\beta^{(k)}\log\rho$ removes the fitted linear association between log score magnitude and log gradient norm on the fitting sample. It does not imply statistical independence or identify a causal effect of norm. Table~\ref{tab:beta_values} reports the fitted slopes. On Olmo3, they range from 2.00 to 2.30 across tasks, motivating task-specific rather than fixed norm correction.

\begin{table}[h]
    \centering
    \caption{Power-law slopes $\beta$ fitted in log space, by model, validation task, and response source.}
    \label{tab:beta_values}
    \begin{tabular}{lcccc}
        \toprule
        \multirow{2}{*}{\textbf{Validation Task}} & \multicolumn{2}{c}{\textbf{Olmo3-7B-Instruct-SFT}} & \multicolumn{2}{c}{\textbf{OpenR1-Distill-7B}} \\
        \cmidrule(lr){2-3} \cmidrule(lr){4-5}
        & \textbf{Off-Policy} & \textbf{On-Policy} & \textbf{Off-Policy} & \textbf{On-Policy} \\
        \midrule
        MATH & 1.59 & 2.21 & 1.17 & 0.92 \\
        BBH & 1.51 & 2.05 & 1.22 & 1.24 \\
        MBPP & 1.47 & 2.30 & 1.14 & 0.98 \\
        MMLU-Pro & 1.48 & 2.00 & 1.22 & 1.16 \\
        \bottomrule
    \end{tabular}
\end{table}

The correction preserves the sign of each task's raw score. With task-specific slopes, the combined score can be written as

\[
S(g)=g^\top\sum_k\rho^{-\beta^{(k)}}c_k.
\]

The combined direction therefore depends on the candidate example's norm. The final score is a ranking transformation, not, in general, an inner product with a single fixed validation gradient.

\section{Limitations}
\label{sec:limitations}

\paragraph{Refinement as a Proxy for Distribution Quality.}
Our broader objective is to improve the performance attainable through SFT, but our experiments evaluate distributions by continuing training from a completed checkpoint. We do not repeat full SFT under each distribution. A gain in this protocol does not establish the same gain when the distribution is used throughout SFT, nor does it identify an optimal distribution for a full training run.

\paragraph{Scope of Influence Approximations.}
We evaluate DRIFT with GraSS sparse projections. Whether on-policy validation provides similar gains with structured Hessian approximations such as EK-FAC, or with dense first-order gradient products, remains untested.

\paragraph{Theoretical Origins of Gradient Norm Bias.}
The relationship between raw score magnitude and gradient norm differs across tasks. Our correction removes a fitted linear association in log space, and its value is assessed through downstream performance. We do not yet explain why the slopes differ or when the correction will improve selection on a new task.

\paragraph{Scope of the Theoretical Analysis.}
Proposition~\ref{thm:on_policy_shift} assumes exact model sampling and full-sequence log-likelihood gradients. The implemented decoder and token-mean loss instead define the surrogate objective $\mathcal{J}_{PG}$ in Eq.~(\ref{eq:pg_if}). The analysis does not bound the error of projected, norm-corrected scores as estimates of retraining effects, the parameter drift during continual SFT, finite-sample variance, or higher-order terms. Figure~\ref{fig:proximity} and Table~\ref{tab:centroid_norms} provide parameter-drift and centroid diagnostics consistent with the locality motivation; they do not establish that the proximity gap is reduced or eliminated.

\paragraph{Model Coverage.}
The main experiments and all seed and sensitivity analyses use one 7B instruction model. The reasoning-focused model in Appendix~\ref{app:openr1} follows the same protocol and again places DRIFT first on the aggregate metrics, but each result is a single run. Larger models and broader coverage of model families remain untested, so the extent to which the gains transfer to those settings is unknown.

\paragraph{Completeness of Anomaly Filtering.}
Per-task correction lowers the ranks of the examples in Tables~\ref{tab:pathological} and \ref{tab:pathological_3}, but some malformed or noisy responses still receive high scores. Gradient norm alone is not a general criterion for identifying such examples.

\section{More Qualitative Cases}
\label{app:qualitative_cases}

Table~\ref{tab:pathological_3} complements the main-text example in Table~\ref{tab:pathological} with two further malformed training examples. Per-task correction places both outside the top-10\% selection cutoff.

\begin{table}[h]
        \caption{Percentile ranks of two additional malformed training examples with large gradient norms under different score corrections.}
        \label{tab:pathological_3}
    \begin{tabular}{@{}p{\linewidth}@{}} 
    
        \toprule
        \textbf{Ill-formed Training Sample} \\
        \midrule
        \textit{\textbf{Prompt:}} Which condiment was used as medicine during the 1830s? \\ 
        \textit{\textbf{Response:}} \color{red}Ketchup.wdgywdufqtufdufwytdfwfdytfwtydvutwqfdytfw \\
        \midrule
        \textbf{Method Rankings (Percentile)} \\
        \begin{tabular}{@{}p{0.45\linewidth} p{0.5\linewidth}@{}} 
            No Debiasing: Top 0.037\% & Packing Correction: Top 0.008\% \\
            Cosine: Top 2.779\% & \textbf{Per-Task Orth.: Top 33.994\%} \\
        \end{tabular} \\
        \bottomrule
    \end{tabular}
    
    \begin{tabular}{@{}p{\linewidth}@{}} 

        \toprule
        \textbf{Ill-formed Training Sample} \\
        \midrule
        \textit{\textbf{Prompt:}} 4: Reer Baniqureeda goortii ay duldageen Saxaabada oo ay u timid Cabsi daran Ninkii la dhihijiray Kacab ibnu Asad oo ahaa Madaxdoodii ayaa wuu u soo jeediyay 3 arimood oo uu yiraahday mid kaqaato sadaxdaas, ee sheeg Sadaxdaas?\\(\textbf{Translation}: When the Quraysh realized that the Companions had arrived in Abyssinia and were under protection, they felt great anxiety. Their leader, Amr ibn al-As, proposed three arguments to the King and asked him to accept one of them. What were those three?) \\ 
        \textit{\textbf{Response:}} \color{red}14: ……………..? \\
        \midrule
        \textbf{Method Rankings (Percentile)} \\
        \begin{tabular}{@{}p{0.45\linewidth} p{0.5\linewidth}@{}} 
            No Debiasing: Top 0.056\% & Packing Correction: Top 7.107\% \\
            Cosine: Top 2.177\% & \textbf{Per-Task Orth.: Top 37.743\%} \\
        \end{tabular} \\
        \bottomrule
    \end{tabular}
\end{table}

\section{Extended Related Work}
\label{app:extended_relatedworks}

\paragraph{Data curation for supervised fine-tuning.}
Data-scoring methods differ in the signals used to value an example. Heuristic filters rely on text-level rules \citep{wenzek2020ccnet}, whereas model-guided filters use learned quality or difficulty estimates \citep{li2024superfiltering,wettig2024qurating}. Meta-raters and sample-level mixing methods learn scores or weights for pretraining \citep{zhuang2025metarater,xi2025samplemix}. Other approaches combine influence with cluster-level diversity \citep{zhang2025quad}, use curvature-aware influence for CLIP adaptation \citep{zhuang2025chips}, or select subsets for instruction tuning \citep{xia2024less,zhang2025d3,humane2025influence}.

Several methods define data value through the model's current learning needs. RHO-LOSS ranks examples by the loss gap between the current model and a reference model trained on held-out data \citep{mindermann2022rholoss}; Rho-1 applies this idea at the token level \citep{lin2024rho1}. DRIFT likewise uses a model-dependent signal, but defines it through the correctness-weighted on-policy PG objective $\mathcal{J}_{PG}$. It then scores corpus examples by their estimated effect on that objective.

Generated-response methods make a different use of the model's outputs. Rejection sampling, self-distillation, and on-policy distillation train directly on generated text \citep{singh2023beyond,xiong2025minimalist,agarwal2024gkd,lu2025onpolicy}. In DRIFT, generated responses define $\mathcal{J}_{PG}$ for selecting examples from the original corpus, and continued training minimizes $\mathcal{L}_{CE}$ on those examples. The validation queries can therefore guide selection beyond their own response set.

\paragraph{Influence functions and data attribution.}
Influence functions estimate how individual training examples affect a model's validation loss \citep{koh2017understanding}. Practical approximations for neural networks include structured Hessian estimates \citep{grosse2023studying,kwon2023datainf,coalson2025if}, random projections \citep{park2023trak,hu2025grass}, and Hessian-free gradient products \citep{xia2024less}.

Gradient-based selection predates LLM instruction tuning. GLISTER and GradMatch select subsets whose aggregate gradients match a validation gradient \citep{killamsetty2021glister,killamsetty2021gradmatch}, and GREATS applies a related first-order criterion online to LLM training batches \citep{wang2024greats}. TRAK \citep{park2023trak} and DataModels \citep{ilyas2022datamodels} predict model behavior from training data. DsDm and MATES estimate target-task effects using datamodels or a learned influence proxy \citep{engstrom2024dsdm,yu2024mates}. CROPI \citep{zhu2025data}, discussed in Section~\ref{sec:related_work}, is particularly relevant because it uses reward-signed policy gradients to select RLVR prompts.

Within this literature, DRIFT addresses two choices in influence-based selection. Its on-policy PG formulation determines the validation responses and objective, motivated by locality concerns in influence approximation \citep{li2024influence,bae2022if,epifano2023revisiting}. Its task-specific correction addresses the dependence of raw scores on gradient magnitude \citep{hu2025grass,xia2024less}. Both choices serve the goal of improving final model performance, evaluated here through checkpoint refinement.

%% file: table/valset.tex
\begin{table}[h]
    \centering
    \caption{Validation queries and on-policy rollouts for Olmo3-7B-Instruct-SFT, with 32 rollouts per query. The final two columns report the proportions of queries for which all sampled rollouts are correct or all are incorrect.}
    \label{tab:valset}
    \small
    \begin{tabular}{l r r r r r}
        \toprule
        \textbf{Task} & \textbf{Queries} & \textbf{Rollouts} & \textbf{Accuracy} & \textbf{All correct} & \textbf{All incorrect} \\
        \midrule
        MATH & 7,500 & 240,000 & 49.1\% & 0.01\% & 10.0\% \\
        MBPP & 596 & 19,072 & 31.1\% & 0.3\% & 38.9\% \\
        MMLU-Pro & 70 & 2,240 & 32.0\% & 10.0\% & 58.6\% \\
        BBH & 54 & 1,728 & 48.6\% & 3.7\% & 27.8\% \\
        \midrule
        Total & 8,220 & 263,040 & 47.6\% & 0.15\% & 12.7\% \\
        \bottomrule
    \end{tabular}
\end{table}

%% file: table/table_robustness_olmo3.tex
\begin{table}[h]
    \centering
    \caption{Olmo3-7B-Instruct-SFT results over three continual-training seeds, reported as mean $\pm$ standard deviation. Each method uses the same selected subset across seeds. Sampling-based benchmarks also average over multiple decoding runs (Appendix~\ref{app:setup}), so the deviations reflect both training and evaluation randomness. Table~\ref{tab:merged_models} reports the means for corpus selectors. S.D. denotes Self-Distillation; Rand. denotes Random.}
    \label{tab:robustness_olmo3}
    \providecommand{\pmstd}[1]{{\scriptsize$\pm$#1}}
\resizebox{\textwidth}{!}{
        \begin{tabular}{l | cccc | cccc | ccc}
            \toprule
            & \multicolumn{4}{c|}{\textbf{Validation-Source Benchmarks}} & \multicolumn{4}{c|}{\textbf{Additional Benchmarks}} & \multicolumn{3}{c}{\textbf{Average}} \\
            \cmidrule(lr){2-5} \cmidrule(lr){6-9} \cmidrule(lr){10-12}
            \textbf{$\uparrow$Acc.(\%)} & \textbf{BBH} & \textbf{MBPP+} & \textbf{MATH} & \textbf{MMLU-Pro} & \textbf{ZebraLogic} & \textbf{LCB v5} & \textbf{OlympiadBench} & \textbf{GPQA-D} & \textbf{Val.-src.} & \textbf{Additional} & \textbf{All} \\
            \midrule
            \textbf{Olmo3-7B-Instruct-SFT} & 44.63 & 56.42 & 66.16 & 44.22 & 17.23 & 15.43 & 34.27 & 35.01 & 52.86 & 25.49 & 39.17 \\
            \midrule
            Rand. & 44.45\pmstd{0.29} & 56.34\pmstd{0.17} & 66.40\pmstd{0.28} & 44.25\pmstd{0.15} & 17.67\pmstd{0.96} & 14.38\pmstd{0.63} & 35.04\pmstd{0.47} & 33.77\pmstd{0.68} & 52.86\pmstd{0.14} & 25.21\pmstd{0.16} & 39.04\pmstd{0.15} \\
            S.D. & 43.18\pmstd{0.68} & 54.77\pmstd{0.41} & 67.90\pmstd{0.36} & 41.28\pmstd{0.33} & 17.76\pmstd{0.23} & 8.92\pmstd{0.52} & 35.99\pmstd{0.38} & 31.29\pmstd{0.31} & 51.78\pmstd{0.17} & 23.49\pmstd{0.11} & 37.64\pmstd{0.13} \\
            S.D. + Rand. & 43.73\pmstd{0.75} & 55.95\pmstd{0.26} & 67.91\pmstd{0.27} & 42.64\pmstd{0.45} & 17.85\pmstd{0.22} & 9.68\pmstd{0.43} & 36.20\pmstd{0.30} & 33.00\pmstd{0.97} & 52.56\pmstd{0.27} & 24.18\pmstd{0.24} & 38.37\pmstd{0.22} \\
            DSIR & 39.97\pmstd{0.23} & 52.63\pmstd{0.88} & 55.11\pmstd{0.42} & 36.22\pmstd{0.82} & 17.68\pmstd{0.34} & 12.73\pmstd{0.68} & 27.83\pmstd{1.07} & 32.82\pmstd{0.70} & 45.98\pmstd{0.09} & 22.76\pmstd{0.04} & 34.37\pmstd{0.02} \\
            Qurating & 44.27\pmstd{0.21} & 55.96\pmstd{0.05} & 66.89\pmstd{0.37} & 44.14\pmstd{0.16} & 17.77\pmstd{0.58} & 15.49\pmstd{0.94} & 34.98\pmstd{0.26} & 34.61\pmstd{0.55} & 52.82\pmstd{0.17} & 25.71\pmstd{0.31} & 39.26\pmstd{0.23} \\
            BM25 & 44.44\pmstd{0.40} & 55.74\pmstd{0.99} & 67.82\pmstd{0.27} & 44.66\pmstd{0.14} & 17.48\pmstd{0.51} & 15.13\pmstd{1.04} & 36.18\pmstd{0.64} & 34.90\pmstd{1.36} & 53.16\pmstd{0.12} & 25.92\pmstd{0.42} & 39.54\pmstd{0.19} \\
            LESS & 43.38\pmstd{0.21} & 53.80\pmstd{0.20} & 64.60\pmstd{0.40} & 43.38\pmstd{0.14} & 16.14\pmstd{0.90} & 13.45\pmstd{0.70} & 33.06\pmstd{0.63} & 33.21\pmstd{0.51} & 51.29\pmstd{0.20} & 23.96\pmstd{0.40} & 37.63\pmstd{0.29} \\
            RDS & 43.68\pmstd{0.13} & 55.59\pmstd{0.24} & 66.56\pmstd{0.39} & 44.80\pmstd{0.18} & 17.76\pmstd{1.09} & 14.59\pmstd{0.84} & 35.38\pmstd{0.55} & 34.27\pmstd{0.67} & 52.66\pmstd{0.21} & 25.50\pmstd{0.29} & 39.08\pmstd{0.24} \\
            IF~(GraSS) & 44.58\pmstd{0.35} & 54.97\pmstd{0.28} & 65.69\pmstd{0.26} & 44.81\pmstd{0.21} & 16.98\pmstd{0.35} & 13.67\pmstd{0.78} & 34.21\pmstd{0.17} & 34.35\pmstd{0.41} & 52.51\pmstd{0.21} & 24.80\pmstd{0.17} & 38.66\pmstd{0.08} \\
            \textbf{\mymethod~(Ours)} & 45.86\pmstd{0.30} & 56.40\pmstd{0.22} & 67.84\pmstd{0.43} & 44.94\pmstd{0.19} & 19.95\pmstd{0.94} & 16.28\pmstd{0.58} & 35.25\pmstd{0.56} & 34.57\pmstd{0.92} & 53.76\pmstd{0.13} & 26.51\pmstd{0.24} & 40.14\pmstd{0.12} \\
            \bottomrule
        \end{tabular}
    }
\end{table}

%% file: table/table_weighting_seeds.tex
\begin{table}[h]
    \centering
    \caption{Per-benchmark results for the validation-objective and rollout-budget ablations in Table~\ref{tab:ablation_weighting}, reported as mean $\pm$ standard deviation over three continual-training seeds on Olmo3-7B-Instruct-SFT. Off-policy uses reference-based CE. The remaining rows use the indicated numbers of positive (P) and negative (N) on-policy responses, with incorrect-response weight $r_{\mathrm{neg}}$. All configurations use per-task orthogonalization. The 32-rollout PG configuration is the DRIFT result in Table~\ref{tab:robustness_olmo3}.}
    \label{tab:weighting_seeds}
    \providecommand{\pmstd}[1]{{\scriptsize$\pm$#1}}
\resizebox{\textwidth}{!}{
        \begin{tabular}{l | cccc | cccc | ccc}
            \toprule
            & \multicolumn{4}{c|}{\textbf{Validation-Source Benchmarks}} & \multicolumn{4}{c|}{\textbf{Additional Benchmarks}} & \multicolumn{3}{c}{\textbf{Average}} \\
            \cmidrule(lr){2-5} \cmidrule(lr){6-9} \cmidrule(lr){10-12}
            \textbf{$\uparrow$Acc.(\%)} & \textbf{BBH} & \textbf{MBPP+} & \textbf{MATH} & \textbf{MMLU-Pro} & \textbf{ZebraLogic} & \textbf{LCB v5} & \textbf{OlympiadBench} & \textbf{GPQA-D} & \textbf{Val.-src.} & \textbf{Additional} & \textbf{All} \\
            \midrule
            Off-policy reference & 43.84\pmstd{0.28} & 55.06\pmstd{0.29} & 66.47\pmstd{0.29} & 44.93\pmstd{0.26} & 16.93\pmstd{0.51} & 13.73\pmstd{0.79} & 34.35\pmstd{0.29} & 33.66\pmstd{0.35} & 52.58\pmstd{0.24} & 24.67\pmstd{0.20} & 38.62\pmstd{0.11} \\
            \midrule
            1P & 44.51\pmstd{0.23} & 56.54\pmstd{0.32} & 67.08\pmstd{0.42} & 45.14\pmstd{0.15} & 18.06\pmstd{0.88} & 15.39\pmstd{0.51} & 35.05\pmstd{0.62} & 34.50\pmstd{0.82} & 53.32\pmstd{0.18} & 25.75\pmstd{0.18} & 39.53\pmstd{0.16} \\
            1P + 1N ($r_{\mathrm{neg}}=-1$) & 44.92\pmstd{0.34} & 56.63\pmstd{0.74} & 67.56\pmstd{0.19} & 45.66\pmstd{0.28} & 19.46\pmstd{0.96} & 15.85\pmstd{0.75} & 34.73\pmstd{0.60} & 34.96\pmstd{0.38} & 53.69\pmstd{0.27} & 26.25\pmstd{0.26} & 39.97\pmstd{0.25} \\
            All P & 45.24\pmstd{0.17} & 56.89\pmstd{0.56} & 67.90\pmstd{0.46} & 43.84\pmstd{0.22} & 17.73\pmstd{0.69} & 13.22\pmstd{0.51} & 36.34\pmstd{0.13} & 35.04\pmstd{1.72} & 53.47\pmstd{0.19} & 25.58\pmstd{0.33} & 39.53\pmstd{0.21} \\
            \midrule
            All P + All N, 4 rollouts & 45.17\pmstd{0.17} & 56.73\pmstd{0.48} & 67.72\pmstd{0.40} & 45.35\pmstd{0.19} & 18.62\pmstd{0.52} & 15.80\pmstd{0.67} & 35.07\pmstd{0.27} & 34.54\pmstd{0.47} & 53.74\pmstd{0.14} & 26.01\pmstd{0.13} & 39.87\pmstd{0.13} \\
            All P + All N, 8 rollouts & 45.61\pmstd{0.08} & 56.02\pmstd{0.55} & 67.56\pmstd{0.36} & 45.18\pmstd{0.30} & 19.71\pmstd{0.79} & 16.01\pmstd{0.76} & 35.73\pmstd{0.50} & 34.12\pmstd{0.56} & 53.59\pmstd{0.10} & 26.40\pmstd{0.10} & 39.99\pmstd{0.02} \\
            All P + All N, 16 rollouts & 45.47\pmstd{0.16} & 56.24\pmstd{0.35} & 67.43\pmstd{0.41} & 45.04\pmstd{0.14} & 18.37\pmstd{0.56} & 15.45\pmstd{0.86} & 35.91\pmstd{0.56} & 35.65\pmstd{1.22} & 53.54\pmstd{0.01} & 26.35\pmstd{0.25} & 39.95\pmstd{0.12} \\
            \textbf{All P + All N, 32 rollouts (\mymethod)} & 45.86\pmstd{0.30} & 56.40\pmstd{0.22} & 67.84\pmstd{0.43} & 44.94\pmstd{0.19} & 19.95\pmstd{0.94} & 16.28\pmstd{0.58} & 35.25\pmstd{0.56} & 34.57\pmstd{0.92} & 53.76\pmstd{0.13} & 26.51\pmstd{0.24} & 40.14\pmstd{0.12} \\
            \midrule
            All P + All N, $r_{\mathrm{neg}}=+1$ & 44.02\pmstd{0.12} & 56.81\pmstd{0.45} & 66.85\pmstd{0.14} & 43.80\pmstd{0.30} & 17.17\pmstd{0.91} & 13.26\pmstd{0.60} & 36.02\pmstd{0.47} & 34.31\pmstd{0.36} & 52.87\pmstd{0.21} & 25.19\pmstd{0.16} & 39.03\pmstd{0.18} \\
            \bottomrule
        \end{tabular}
    }
\end{table}

%% file: table/table_debias_seeds.tex
\begin{table}[h]
    \centering
    \caption{Per-benchmark results for the gradient-debiasing ablation in Table~\ref{tab:ablation_debiasing}, reported as mean $\pm$ standard deviation over three continual-training seeds on Olmo3-7B-Instruct-SFT. All variants use the same on-policy PG objective $\mathcal{J}_{PG}$ and differ only in norm correction. Per-Task Orth. is the DRIFT result in Table~\ref{tab:robustness_olmo3}.}
    \label{tab:debias_seeds}
    \providecommand{\pmstd}[1]{{\scriptsize$\pm$#1}}
\resizebox{\textwidth}{!}{
        \begin{tabular}{l | cccc | cccc | ccc}
            \toprule
            & \multicolumn{4}{c|}{\textbf{Validation-Source Benchmarks}} & \multicolumn{4}{c|}{\textbf{Additional Benchmarks}} & \multicolumn{3}{c}{\textbf{Average}} \\
            \cmidrule(lr){2-5} \cmidrule(lr){6-9} \cmidrule(lr){10-12}
            \textbf{$\uparrow$Acc.(\%)} & \textbf{BBH} & \textbf{MBPP+} & \textbf{MATH} & \textbf{MMLU-Pro} & \textbf{ZebraLogic} & \textbf{LCB v5} & \textbf{OlympiadBench} & \textbf{GPQA-D} & \textbf{Val.-src.} & \textbf{Additional} & \textbf{All} \\
            \midrule
            No Debiasing & 43.65\pmstd{0.20} & 55.94\pmstd{1.01} & 65.47\pmstd{0.17} & 44.61\pmstd{0.06} & 19.55\pmstd{0.65} & 16.10\pmstd{0.71} & 34.67\pmstd{0.28} & 35.44\pmstd{0.40} & 52.42\pmstd{0.23} & 26.44\pmstd{0.08} & 39.43\pmstd{0.15} \\
            Cosine & 44.70\pmstd{0.34} & 56.32\pmstd{0.35} & 67.11\pmstd{0.35} & 45.12\pmstd{0.29} & 20.86\pmstd{0.70} & 15.53\pmstd{0.67} & 35.92\pmstd{0.21} & 33.81\pmstd{0.35} & 53.31\pmstd{0.18} & 26.53\pmstd{0.12} & 39.92\pmstd{0.06} \\
            Packing Correction & 44.01\pmstd{0.43} & 56.07\pmstd{0.39} & 66.61\pmstd{0.17} & 44.79\pmstd{0.27} & 20.87\pmstd{1.10} & 16.52\pmstd{0.77} & 35.24\pmstd{0.39} & 34.04\pmstd{0.16} & 52.87\pmstd{0.30} & 26.67\pmstd{0.04} & 39.77\pmstd{0.17} \\
            Unified Orth. & 44.98\pmstd{0.41} & 56.73\pmstd{0.62} & 67.53\pmstd{0.40} & 45.20\pmstd{0.05} & 19.39\pmstd{0.69} & 15.33\pmstd{0.85} & 36.58\pmstd{0.31} & 33.72\pmstd{0.58} & 53.61\pmstd{0.15} & 26.26\pmstd{0.10} & 39.93\pmstd{0.04} \\
            \textbf{Per-Task Orth. (\mymethod)} & 45.86\pmstd{0.30} & 56.40\pmstd{0.22} & 67.84\pmstd{0.43} & 44.94\pmstd{0.19} & 19.95\pmstd{0.94} & 16.28\pmstd{0.58} & 35.25\pmstd{0.56} & 34.57\pmstd{0.92} & 53.76\pmstd{0.13} & 26.51\pmstd{0.24} & 40.14\pmstd{0.12} \\
            \bottomrule
        \end{tabular}
    }
\end{table}

%% file: table/table_scoring_seeds.tex
\begin{table}[htbp]
    \centering
    \caption{Per-benchmark scoring ablation on Olmo3-7B-Instruct-SFT, reported as mean $\pm$ standard deviation over three continual-training seeds. BM25 (signed rollouts) uses on-policy responses with signed correctness weighting. BM25 and DRIFT results are repeated from Table~\ref{tab:robustness_olmo3}.}
    \label{tab:scoring_seeds}
    \small
    \providecommand{\pmstd}[1]{{\scriptsize$\pm$#1}}
    \begin{tabular}{lccc}
        \toprule
        \textbf{Benchmark} & \textbf{BM25} & \textbf{BM25 (signed rollouts)} & \textbf{\mymethod} \\
        \midrule
        BBH & 44.44\pmstd{0.40} & 44.73\pmstd{0.20} & 45.86\pmstd{0.30} \\
        MBPP+ & 55.74\pmstd{0.99} & 56.50\pmstd{0.23} & 56.40\pmstd{0.22} \\
        MATH & 67.82\pmstd{0.27} & 66.92\pmstd{0.36} & 67.84\pmstd{0.43} \\
        MMLU-Pro & 44.66\pmstd{0.14} & 43.85\pmstd{0.27} & 44.94\pmstd{0.19} \\
        \midrule
        ZebraLogic & 17.48\pmstd{0.51} & 17.54\pmstd{0.17} & 19.95\pmstd{0.94} \\
        LCB v5 & 15.13\pmstd{1.04} & 14.40\pmstd{0.19} & 16.28\pmstd{0.58} \\
        OlympiadBench & 36.18\pmstd{0.64} & 35.63\pmstd{0.36} & 35.25\pmstd{0.56} \\
        GPQA-D & 34.90\pmstd{1.36} & 34.84\pmstd{1.39} & 34.57\pmstd{0.92} \\
        \midrule
        Validation-source average & 53.16\pmstd{0.12} & 53.00\pmstd{0.10} & 53.76\pmstd{0.13} \\
        Additional average & 25.92\pmstd{0.42} & 25.60\pmstd{0.39} & 26.51\pmstd{0.24} \\
        All average & 39.54\pmstd{0.19} & 39.30\pmstd{0.15} & 40.14\pmstd{0.12} \\
        \bottomrule
    \end{tabular}
\end{table}

%% file: table/cost.tex
\begin{table}[h]
    \centering
    \caption{Wall-clock costs for Olmo3-7B-Instruct-SFT, in hours on one node with eight A800 GPUs. Corpus preparation is reusable; projected features require the checkpoint, corpus, parameter subset, and projection configuration to remain fixed. Validation-stage work is repeated when queries or response construction change. A dagger ($\dagger$) marks work that is reusable in principle but was recomputed here; ``est.'' marks estimates. Generation times of one and two hours, and totals based on them, are linearly scaled estimates. Totals containing stages shorter than 0.5 hours are rounded upper approximations. Rows below the rule compare approximately matched elapsed-time budgets (Section~\ref{sec:ratio_sweep}). S.D. denotes Self-Distillation; Rand. denotes Random.}
    \label{tab:cost}
    \resizebox{\textwidth}{!}{%
    \begin{tabular}{l r r r r r r}
        \toprule
        \textbf{Method} & \textbf{Per corpus} & \textbf{Per val. set} & \textbf{Continual SFT} & \textbf{Total (h)} & \textbf{Storage} & \textbf{All} \\
        \midrule
        Rand. & -- & -- & 6.6 & 6.6 & -- & 39.04 \\
        BM25 & 0.2$^\dagger$ & 4.0 & 6.6 & 10.8 & -- & 39.54 \\
        DSIR & 0.8$^\dagger$ & 0.9 & 6.6 & 8.3 & -- & 34.37 \\
        Qurating & $\approx$1 (est.) & -- & 6.6 & 7.6 & -- & 39.26 \\
        RDS & 8--11 & 10.9 & 6.6 & 26--29 & -- & 39.08 \\
        S.D. & -- & 8 & 6.6 & 14.6 & -- & 37.64 \\
        S.D. + Rand. & -- & 8 & 6.6 & 14.6 & -- & 38.37 \\
        IF (GraSS) & 12 & $<$0.5 & 6.6 & 19.1 & 63 GB & 38.66 \\
        LESS & 6.6 (warm-up, est.) + 12 & $<$0.5 & 6.6 & 25.7 & 63 GB & 37.63 \\
        \textbf{\mymethod} (32 rollouts) & 12 & 8 + $<$0.5 & 6.6 & 27.1 & 63 GB & \textbf{40.14} \\
        \midrule
        \mymethod, 8 rollouts & 12 & 2 + $<$0.5 & 6.6 & 21.1 & 63 GB & 39.99 \\
        \mymethod, 4 rollouts & 12 & 1 + $<$0.5 & 6.6 & 20.1 & 63 GB & 39.87 \\
        BM25, 25\% data, 1,250 steps & 0.2$^\dagger$ & 4.0 & 16.5 & 20.7 & -- & 39.66 \\
        \bottomrule
    \end{tabular}%
    }
\end{table}

%% file: table/openr1_results.tex
\begin{table}[h]
    \centering
    \caption{Per-benchmark continual-SFT results on OpenR1-Distill-7B, including the self-distillation control. The aggregate results for corpus selectors are also reported in Table~\ref{tab:merged_models}. Each continual-training entry is a single run. Bold indicates the best continual-training result; colored differences are relative to the starting checkpoint. S.D. denotes Self-Distillation; Rand. denotes Random.}
    \label{tab:openr1_results}
\resizebox{\textwidth}{!}{
        \begin{tabular}{l | cccc | cccc | cc}
            \toprule
            & \multicolumn{4}{c|}{\textbf{Validation-Source Benchmarks}} & \multicolumn{4}{c|}{\textbf{Additional Benchmarks}} & \multicolumn{2}{c}{\textbf{Average}} \\
            \cmidrule(lr){2-5} \cmidrule(lr){6-9} \cmidrule(lr){10-11}
            \makecell[c]{\textbf{$\uparrow$Acc.(\%)}} & \makecell[c]{\textbf{BBH} \\ } & \makecell[c]{\textbf{MBPP+} \\ } & \makecell[c]{\textbf{MATH} \\ } & \makecell[c]{\textbf{MMLU} \\ \textbf{Pro}} & \makecell[c]{\textbf{Zebra} \\ \textbf{Logic}} & \makecell[c]{\textbf{LCB} \\ \textbf{v5}} & \makecell[c]{\textbf{Olympiad} \\ \textbf{Bench} } & \makecell[c]{\textbf{GPQA} \\ \textbf{Diamond}} & \makecell[c]{\textbf{All} \\ } & \makecell[c]{\textbf{Additional} \\ } \\
            \midrule
            \textbf{OpenR1-Distill-7B}                  & 76.04 & 54.23 & 90.98 & 51.65 & 23.85 & 43.86 & 67.51 & 49.72 & 57.23& 46.24\\
            \midrule
            \quad + Rand.                     & 76.52 & 53.90 & 91.05 & 51.15 & 23.75 & 44.81 & 66.58 & \textbf{51.45} & 57.40 \textcolor[rgb]{0.62, 0.85, 0.62}{\scriptsize +0.17} & 46.65 \textcolor[rgb]{0.59, 0.84, 0.59}{\scriptsize +0.41} \\
            \quad + S.D.                     & 60.82 & 55.85 & 83.78 & 48.59 & 20.53 & 20.31 & 60.24 & 44.48 & 49.33 \textcolor[rgb]{0.70, 0.00, 0.00}{\scriptsize -7.90} & 36.39 \textcolor[rgb]{0.70, 0.00, 0.00}{\scriptsize -9.85} \\
            \quad + DSIR                           & 73.86 & 54.17 & 90.85 & 49.34 & 22.30 & 44.46 & 65.95 & 49.02 & 56.24 \textcolor[rgb]{0.87, 0.61, 0.61}{\scriptsize -0.99} & 45.43 \textcolor[rgb]{0.88, 0.64, 0.64}{\scriptsize -0.81} \\
            \quad + Qurating                          & \textbf{76.70} & 54.33 & \textbf{91.88} & 51.16 & 24.73 & 47.63 & 68.25 & 47.57 & 57.78 \textcolor[rgb]{0.43, 0.75, 0.43}{\scriptsize +0.55} & 47.05 \textcolor[rgb]{0.47, 0.77, 0.47}{\scriptsize +0.81} \\
            \quad + BM25                           & 75.76 & 54.23 & 91.20 & 51.19 & 24.95 & 44.19 & 67.62 & 49.40 & 57.32 \textcolor[rgb]{0.66, 0.88, 0.66}{\scriptsize +0.09} & 46.54 \textcolor[rgb]{0.62, 0.85, 0.62}{\scriptsize +0.30} \\
            \quad + LESS                          & 76.15 & 54.17 & 91.76 & 50.44 & 24.30 & 44.10 & \textbf{68.32} & 49.02 & 57.28 \textcolor[rgb]{0.68, 0.89, 0.68}{\scriptsize +0.05} & 46.44 \textcolor[rgb]{0.64, 0.87, 0.64}{\scriptsize +0.20} \\
            \quad + RDS                          & 75.88 & 54.86 & 91.00 & \textbf{51.69} & 25.48 & 45.19 & 66.77 & 50.19 & 57.63 \textcolor[rgb]{0.50, 0.79, 0.50}{\scriptsize +0.40} & 46.91 \textcolor[rgb]{0.51, 0.79, 0.51}{\scriptsize +0.67} \\
            \quad + IF~(GraSS)                           & 75.46 & 54.27 & 91.52 & 50.78 & 25.80 & 48.18 & 67.62 & 49.15 & 57.85 \textcolor[rgb]{0.40, 0.73, 0.40}{\scriptsize +0.62} & 47.69 \textcolor[rgb]{0.29, 0.67, 0.29}{\scriptsize +1.45} \\
            \quad + \textbf{\mymethod~(Ours)}   & 76.11   &  \textbf{56.02}     &  90.82     &  51.37     &  \textbf{25.88}     &  \textbf{50.21}     &  68.29     &  50.54     & \textbf{58.66} \textcolor[rgb]{0.00, 0.50, 0.00}{\textbf{\scriptsize +1.43}}      &       \textbf{48.73} \textcolor[rgb]{0.00, 0.50, 0.00}{\textbf{\scriptsize +2.49}}       \\
            \bottomrule
        \end{tabular}
    }
\end{table}

%% file: reference.bib
@article{bae2022if,
  title={If influence functions are the answer, then what is the question?},
  author={Bae, Juhan and Ng, Nathan and Lo, Alston and Ghassemi, Marzyeh and Grosse, Roger B},
  journal={Advances in Neural Information Processing Systems},
  volume={35},
  pages={17953--17967},
  year={2022}
}

@article{xie2023data,
  title={Data selection for language models via importance resampling},
  author={Xie, Sang Michael and Santurkar, Shibani and Ma, Tengyu and Liang, Percy S},
  journal={Advances in Neural Information Processing Systems},
  volume={36},
  pages={34201--34227},
  year={2023}
}

@article{wettig2024qurating,
  title={Qurating: Selecting high-quality data for training language models},
  author={Wettig, Alexander and Gupta, Aatmik and Malik, Saumya and Chen, Danqi},
  journal={arXiv preprint arXiv:2402.09739},
  year={2024}
}

@book{robertson2009probabilistic,
  title={The probabilistic relevance framework: BM25 and beyond},
  author={Robertson, Stephen and Zaragoza, Hugo},
  volume={4},
  year={2009},
  publisher={Now Publishers Inc}
}

@article{xia2024less,
  title={Less: Selecting influential data for targeted instruction tuning},
  author={Xia, Mengzhou and Malladi, Sadhika and Gururangan, Suchin and Arora, Sanjeev and Chen, Danqi},
  journal={arXiv preprint arXiv:2402.04333},
  year={2024}
}

@article{ivison2025large,
  title={Large-scale data selection for instruction tuning},
  author={Ivison, Hamish and Zhang, Muru and Brahman, Faeze and Koh, Pang Wei and Dasigi, Pradeep},
  journal={arXiv preprint arXiv:2503.01807},
  year={2025}
}

@inproceedings{hu2025grass,
  title        = {{GraSS}: Scalable Data Attribution with Gradient Sparsification and Sparse Projection},
  author       = {Hu, Pingbang and Melkonian, Joseph and Tang, Weijing and Zhao, Han and Ma, Jiaqi W.},
  booktitle    = {Advances in Neural Information Processing Systems (NeurIPS 2025)},
  year         = {2025},
  url          = {https://arxiv.org/abs/2505.18976},
  doi          = {10.52202/085713-1464},
  howpublished = {NeurIPS proceedings}
}

@article{xie2023doremi,
  title={Doremi: Optimizing data mixtures speeds up language model pretraining},
  author={Xie, Sang Michael and Pham, Hieu and Dong, Xuanyi and Du, Nan and Liu, Hanxiao and Lu, Yifeng and Liang, Percy S and Le, Quoc V and Ma, Tengyu and Yu, Adams Wei},
  journal={Advances in Neural Information Processing Systems},
  volume={36},
  pages={69798--69818},
  year={2023}
}

@article{liu2024regmix,
  title={Regmix: Data mixture as regression for language model pre-training},
  author={Liu, Qian and Zheng, Xiaosen and Muennighoff, Niklas and Zeng, Guangtao and Dou, Longxu and Pang, Tianyu and Jiang, Jing and Lin, Min},
  journal={arXiv preprint arXiv:2407.01492},
  year={2024}
}

@inproceedings{peng2020domain2vec,
  title={Domain2vec: Domain embedding for unsupervised domain adaptation},
  author={Peng, Xingchao and Li, Yichen and Saenko, Kate},
  booktitle={European conference on computer vision},
  pages={756--774},
  year={2020},
  organization={Springer}
}

@article{wang2024greats,
  title={Greats: Online selection of high-quality data for llm training in every iteration},
  author={Wang, Jiachen T and Wu, Tong and Song, Dawn and Mittal, Prateek and Jia, Ruoxi},
  journal={Advances in Neural Information Processing Systems},
  volume={37},
  pages={131197--131223},
  year={2024}
}

@inproceedings{ouyang2022training,
  title={Training language models to follow instructions with human feedback},
  author={Ouyang, Long and Wu, Jeffrey and Jiang, Xu and Almeida, Diogo and Wainwright, Carroll and Mishkin, Pamela and Zhang, Chong and Agarwal, Sandhini and Slama, Katarina and Ray, Alex and others},
  booktitle={Advances in Neural Information Processing Systems},
  volume={35},
  pages={27730--27744},
  year={2022}
}

@inproceedings{wei2021finetuned,
  title={Finetuned Language Models are Zero-Shot Learners},
  author={Wei, Jason and Bosma, Maarten and Zhao, Vincent and Guu, Kelvin and Yu, Adams Wei and Lester, Brian and Du, Nan and Dai, Andrew M and Le, Quoc V},
  booktitle={International Conference on Learning Representations},
  year={2022}
}

@inproceedings{zhou2023lima,
  title={{LIMA}: Less Is More for Alignment},
  author={Zhou, Chunting and Liu, Pengfei and Xu, Puxin and Iyer, Srinivasan and Sun, Jiao and Mao, Yuning and Ma, Xuezhe and Efrat, Avia and Yu, Ping and Yu, Lili and Zhang, Susan and Ghosh, Gargi and Lewis, Mike and Zettlemoyer, Luke and Levy, Omer},
  booktitle={Advances in Neural Information Processing Systems},
  volume={36},
  year={2023}
}

@inproceedings{ye2024data,
  title={Data Mixing Laws: Optimizing Data Mixtures by Predicting Language Modeling Performance},
  author={Ye, Jiasheng and Liu, Peiju and Sun, Tianxiang and Zhan, Jun and Zhou, Yunhua and Qiu, Xipeng},
  booktitle={The Twelfth International Conference on Learning Representations},
  year={2024}
}

@inproceedings{pruthi2020estimating,
  title={Estimating Training Data Influence by Tracing Gradient Descent},
  author={Pruthi, Garima and Liu, Frederick and Kale, Satyen and Sundararajan, Mukund},
  booktitle={Advances in Neural Information Processing Systems},
  volume={33},
  pages={19920--19930},
  year={2020}
}

@inproceedings{ilyas2022datamodels,
  title={Datamodels: Predicting Predictions from Training Data},
  author={Ilyas, Andrew and Park, Sung Min and Engstrom, Logan and Leclerc, Guillaume and Madry, Aleksander},
  booktitle={International Conference on Machine Learning},
  pages={9825--9840},
  year={2022},
  organization={PMLR}
}

@inproceedings{koh2017understanding,
  title={Understanding Black-box Predictions via Influence Functions},
  author={Koh, Pang Wei and Liang, Percy},
  booktitle={International Conference on Machine Learning},
  pages={1885--1894},
  year={2017},
  organization={PMLR}
}

@article{grosse2023studying,
  title={Studying Large Language Model Generalization with Influence Functions},
  author={Grosse, Roger and Bae, Juhan and Anil, Cem and Elhage, Nelson and Tamkin, Alex and Tajdini, Amirhossein and Steiner, Benoit and Li, Dustin and Durmus, Esin and Perez, Ethan and others},
  journal={arXiv preprint arXiv:2308.03296},
  year={2023}
}

@inproceedings{wenzek2020ccnet,
  title={CCNet: Extracting high quality monolingual datasets from web crawl data},
  author={Wenzek, Guillaume and Lachaux, Marie-Anne and Conneau, Alexis and Chaudhary, Vishrav and Guzm{\'a}n, Francisco and Joulin, Armand and Grave, Edouard},
  booktitle={Proceedings of the twelfth language resources and evaluation conference},
  pages={4003--4012},
  year={2020}
}

@inproceedings{li2024superfiltering,
  title={Superfiltering: Weak-to-strong data filtering for fast instruction-tuning},
  author={Li, Ming and Zhang, Yong and He, Shwai and Li, Zhitao and Zhao, Hongyu and Wang, Jianzong and Cheng, Ning and Zhou, Tianyi},
  booktitle={Proceedings of the 62nd Annual Meeting of the Association for Computational Linguistics (Volume 1: Long Papers)},
  pages={14255--14273},
  year={2024}
}

@article{zhang2025d3,
  title={D3: Diversity, difficulty, and dependability-aware data selection for sample-efficient llm instruction tuning},
  author={Zhang, Jia and Zhang, Chen-Xi and Liu, Yao and Jin, Yi-Xuan and Yang, Xiao-Wen and Zheng, Bo and Liu, Yi and Guo, Lan-Zhe},
  journal={arXiv preprint arXiv:2503.11441},
  year={2025}
}

@article{park2023trak,
  title={Trak: Attributing model behavior at scale},
  author={Park, Sung Min and Georgiev, Kristian and Ilyas, Andrew and Leclerc, Guillaume and Madry, Aleksander},
  journal={arXiv preprint arXiv:2303.14186},
  year={2023}
}

@article{kwon2023datainf,
  title={Datainf: Efficiently estimating data influence in lora-tuned llms and diffusion models},
  author={Kwon, Yongchan and Wu, Eric and Wu, Kevin and Zou, James},
  journal={arXiv preprint arXiv:2310.00902},
  year={2023}
}

@article{li2024influence,
  title={Do influence functions work on large language models},
  author={Li, Zhe and Zhao, Wei and Li, Yige and Sun, Jun},
  journal={arXiv preprint arXiv:2409.19998},
  volume={3},
  year={2024}
}

@article{epifano2023revisiting,
  title={Revisiting the fragility of influence functions},
  author={Epifano, Jacob R and Ramachandran, Ravi P and Masino, Aaron J and Rasool, Ghulam},
  journal={Neural Networks},
  volume={162},
  pages={581--588},
  year={2023},
  publisher={Elsevier}
}

@article{zhu2025data,
  title={Data-Efficient RLVR via Off-Policy Influence Guidance},
  author={Zhu, Erle and Jiang, Dazhi and Wang, Yuan and Li, Xujun and Cheng, Jiale and Gu, Yuxian and Niu, Yilin and Zeng, Aohan and Tang, Jie and Huang, Minlie and others},
  journal={arXiv preprint arXiv:2510.26491},
  year={2025}
}

@incollection{william1984extensions,
  title={Extensions of {L}ipschitz mappings into a {H}ilbert space},
  author={Johnson, William B. and Lindenstrauss, Joram},
  booktitle={Contemporary Mathematics},
  volume={26},
  pages={189--206},
  year={1984},
  publisher={American Mathematical Society}
}

@article{suzgun2022challenging,
  title={Challenging BIG-Bench Tasks and Whether Chain-of-Thought Can Solve Them},
  author={Suzgun, Mirac and Scales, Nathan and Sch{\"a}rli, Nathanael and Gehrmann, Sebastian and Tay, Yi and Chung, Hyung Won and Chowdhery, Aakanksha and Le, Quoc V and Chi, Ed H and Zhou, Denny and Wei, Jason},
  journal={arXiv preprint arXiv:2210.09261},
  year={2022}
}

@inproceedings{evalplus,
  title = {Is Your Code Generated by Chat{GPT} Really Correct? Rigorous Evaluation of Large Language Models for Code Generation},
  author = {Liu, Jiawei and Xia, Chunqiu Steven and Wang, Yuyao and Zhang, Lingming},
  booktitle = {Thirty-seventh Conference on Neural Information Processing Systems},
  year = {2023},
  url = {https://openreview.net/forum?id=1qvx610Cu7},
}

@article{wang2024mmlu,
  title={Mmlu-pro: A more robust and challenging multi-task language understanding benchmark},
  author={Wang, Yubo and Ma, Xueguang and Zhang, Ge and Ni, Yuansheng and Chandra, Abhranil and Guo, Shiguang and Ren, Weiming and Arulraj, Aaran and He, Xuan and Jiang, Ziyan and others},
  journal={arXiv preprint arXiv:2406.01574},
  year={2024}
}

@misc{zebralogic2024,
    title={ZebraLogic: Benchmarking the Logical Reasoning Ability of Language Models},
    author={Bill Yuchen Lin and Ronan Le Bras and Yejin Choi},
    url={https://huggingface.co/spaces/allenai/ZebraLogic},
    year={2024}
}

@article{jain2024livecodebench,
  title={Livecodebench: Holistic and contamination free evaluation of large language models for code},
  author={Jain, Naman and Han, King and Gu, Alex and Li, Wen-Ding and Yan, Fanjia and Zhang, Tianjun and Wang, Sida and Solar-Lezama, Armando and Sen, Koushik and Stoica, Ion},
  journal={arXiv preprint arXiv:2403.07974},
  year={2024}
}

@inproceedings{he2024olympiadbench,
  title={Olympiadbench: A challenging benchmark for promoting agi with olympiad-level bilingual multimodal scientific problems},
  author={He, Chaoqun and Luo, Renjie and Bai, Yuzhuo and Hu, Shengding and Thai, Zhen and Shen, Junhao and Hu, Jinyi and Han, Xu and Huang, Yujie and Zhang, Yuxiang and others},
  booktitle={Proceedings of the 62nd Annual Meeting of the Association for Computational Linguistics (Volume 1: Long Papers)},
  pages={3828--3850},
  year={2024}
}

@article{rein2023gpqa,
  title={Gpqa: A graduate-level google-proof q\&a benchmark},
  author={Rein, David and Hou, Betty Li and Stickland, Asa Cooper and Petty, Jackson and Pang, Richard Yuanzhe and Dirani, Julien and Michael, Julian and Bowman, Samuel R},
  journal={arXiv preprint arXiv:2311.12022},
  year={2023}
}

@article{mukherjee2025reinforcement,
  title={Reinforcement learning finetunes small subnetworks in large language models},
  author={Mukherjee, Sagnik and Yuan, Lifan and Hakkani-Tur, Dilek and Peng, Hao},
  journal={arXiv preprint arXiv:2505.11711},
  year={2025}
}

@article{choe2024your,
  title={What is your data worth to gpt? llm-scale data valuation with influence functions},
  author={Choe, Sang Keun and Ahn, Hwijeen and Bae, Juhan and Zhao, Kewen and Kang, Minsoo and Chung, Youngseog and Pratapa, Adithya and Neiswanger, Willie and Strubell, Emma and Mitamura, Teruko and others},
  journal={arXiv preprint arXiv:2405.13954},
  year={2024}
}

@article{lu2024bm25s,
  title={Bm25s: Orders of magnitude faster lexical search via eager sparse scoring},
  author={L{\`u}, Xing Han},
  journal={arXiv preprint arXiv:2407.03618},
  year={2024}
}

@misc{singh2023beyond,
  title         = {Beyond Human Data: Scaling Self-Training for Problem-Solving with Language Models},
  author        = {Singh, Avi and Co-Reyes, John D. and Agarwal, Rishabh and Anand, Ankesh and Patil, Piyush and others},
  year          = {2023},
  eprint        = {2312.06585},
  archivePrefix = {arXiv},
  primaryClass  = {cs.LG},
  doi           = {10.48550/arXiv.2312.06585},
  url           = {https://arxiv.org/abs/2312.06585}
}

@article{xiong2025minimalist,
  title={A minimalist approach to llm reasoning: from rejection sampling to reinforce},
  author={Xiong, Wei and Yao, Jiarui and Xu, Yuhui and Pang, Bo and Wang, Lei and Sahoo, Doyen and Li, Junnan and Jiang, Nan and Zhang, Tong and Xiong, Caiming and others},
  journal={arXiv preprint arXiv:2504.11343},
  year={2025}
}

@article{olmo2025olmo,
  title={Olmo 3},
  author={Olmo, Team and Ettinger, Allyson and Bertsch, Amanda and Kuehl, Bailey and Graham, David and Heineman, David and Groeneveld, Dirk and Brahman, Faeze and Timbers, Finbarr and Ivison, Hamish and others},
  journal={arXiv preprint arXiv:2512.13961},
  year={2025}
}

@misc{openr1,
    title = {Open R1: A fully open reproduction of DeepSeek-R1},
    url = {https://github.com/huggingface/open-r1},
    author = {HuggingFace},
    month = {January},
    year = {2025}
}

@article{coalson2025if,
  title={If-guide: Influence function-guided detoxification of llms},
  author={Coalson, Zachary and Bae, Juhan and Carlini, Nicholas and Hong, Sanghyun},
  journal={arXiv preprint arXiv:2506.01790},
  year={2025}
}

@article{humane2025influence,
  title={Influence functions for efficient data selection in reasoning},
  author={Humane, Prateek and Cudrano, Paolo and Kaplan, Daniel Z and Matteucci, Matteo and Chakraborty, Supriyo and Rish, Irina},
  journal={arXiv preprint arXiv:2510.06108},
  year={2025}
}

@inproceedings{zhuang2025metarater,
  title={Meta-rater: A Multi-dimensional Data Selection Method for Pre-training Language Models},
  author={Zhuang, Xinlin and Peng, Jiahui and Ma, Ren and Wang, Yinfan and Bai, Tianyi and Wei, Xingjian and Qiu, Jiantao and Zhang, Chi and Qian, Ying and He, Conghui},
  booktitle={Proceedings of the 63rd Annual Meeting of the Association for Computational Linguistics (Volume 1: Long Papers)},
  pages={10856--10896},
  year={2025},
  url={https://aclanthology.org/2025.acl-long.533/}
}

@inproceedings{zhang2025quad,
  title={Harnessing Diversity for Important Data Selection in Pretraining Large Language Models},
  author={Zhang, Chi and Zhong, Huaping and Zhang, Kuan and Chai, Chengliang and Wang, Rui and Zhuang, Xinlin and Bai, Tianyi and Qiu, Jiantao and Cao, Lei and Fan, Ju and Yuan, Ye and Wang, Guoren and He, Conghui},
  booktitle={The Thirteenth International Conference on Learning Representations},
  year={2025},
  url={https://openreview.net/forum?id=bMC1t7eLRc}
}

@article{zhuang2025chips,
  title={{CHIPS}: Efficient {CLIP} Adaptation via Curvature-aware Hybrid Influence-based Data Selection},
  author={Zhuang, Xinlin and Li, Yichen and Liu, Xiwei and Yang, Haolin and Lu, Yifan and Zou, Ziyun and Li, Yulong and Li, Huifa and Chen, Dongliang and Wang, Qinglei and Liu, Weiyang and Qian, Ying and Shi, Jiangming and Razzak, Imran},
  journal={arXiv preprint arXiv:2511.18519},
  year={2025},
  note={To appear in CVPR 2026},
  url={https://arxiv.org/abs/2511.18519}
}

@inproceedings{xi2025samplemix,
  title={{SampleMix}: A Sample-wise Pre-training Data Mixing Strategy by Coordinating Data Quality and Diversity},
  author={Xi, Xiangyu and Kong, Deyang and Yang, Jian and Yang, Jiawei and Chen, Zhengyu and Wang, Wei and Wang, Jingang and Cai, Xunliang and Zhang, Shikun and Ye, Wei},
  booktitle={Findings of the Association for Computational Linguistics: EMNLP 2025},
  pages={13736--13758},
  year={2025},
  doi={10.18653/v1/2025.findings-emnlp.741},
  url={https://aclanthology.org/2025.findings-emnlp.741/}
}

@inproceedings{mindermann2022rholoss,
  title={Prioritized Training on Points that are Learnable, Worth Learning, and Not Yet Learnt},
  author={Mindermann, S{\"o}ren and Brauner, Jan and Razzak, Muhammed and Sharma, Mrinank and Kirsch, Andreas and Xu, Winnie and H{\"o}ltgen, Benedikt and Gomez, Aidan N. and Morisot, Adrien and Farquhar, Sebastian and Gal, Yarin},
  booktitle={International Conference on Machine Learning},
  year={2022},
  url={https://arxiv.org/abs/2206.07137}
}

@inproceedings{lin2024rho1,
  title={Not All Tokens Are What You Need for Pretraining},
  author={Lin, Zhenghao and Gou, Zhibin and Gong, Yeyun and Liu, Xiao and Shen, Yelong and Xu, Ruochen and Lin, Chen and Yang, Yujiu and Jiao, Jian and Duan, Nan and Chen, Weizhu},
  booktitle={Advances in Neural Information Processing Systems},
  volume={37},
  year={2024},
  note={arXiv:2404.07965, ``Rho-1: Not All Tokens Are What You Need''},
  url={https://arxiv.org/abs/2404.07965}
}

@inproceedings{killamsetty2021gradmatch,
  title={{GRAD-MATCH}: Gradient Matching based Data Subset Selection for Efficient Deep Model Training},
  author={Killamsetty, Krishnateja and Sivasubramanian, Durga and Ramakrishnan, Ganesh and De, Abir and Iyer, Rishabh},
  booktitle={International Conference on Machine Learning},
  year={2021},
  url={https://arxiv.org/abs/2103.00123}
}

@inproceedings{killamsetty2021glister,
  title={{GLISTER}: Generalization based Data Subset Selection for Efficient and Robust Learning},
  author={Killamsetty, Krishnateja and Sivasubramanian, Durga and Ramakrishnan, Ganesh and Iyer, Rishabh},
  booktitle={Proceedings of the AAAI Conference on Artificial Intelligence},
  volume={35},
  pages={8110--8118},
  year={2021},
  url={https://arxiv.org/abs/2012.10630}
}

@inproceedings{yu2024mates,
  title={{MATES}: Model-Aware Data Selection for Efficient Pretraining with Data Influence Models},
  author={Yu, Zichun and Das, Spandan and Xiong, Chenyan},
  booktitle={Advances in Neural Information Processing Systems},
  volume={37},
  year={2024},
  url={https://arxiv.org/abs/2406.06046}
}

@inproceedings{engstrom2024dsdm,
  title={{DsDm}: Model-Aware Dataset Selection with Datamodels},
  author={Engstrom, Logan and Feldmann, Axel and Madry, Aleksander},
  booktitle={International Conference on Machine Learning},
  pages={12491--12526},
  year={2024},
  url={https://proceedings.mlr.press/v235/engstrom24a.html}
}

@inproceedings{agarwal2024gkd,
  title={On-Policy Distillation of Language Models: Learning from Self-Generated Mistakes},
  author={Agarwal, Rishabh and Vieillard, Nino and Zhou, Yongchao and Stanczyk, Piotr and Ramos, Sabela and Geist, Matthieu and Bachem, Olivier},
  booktitle={International Conference on Learning Representations},
  year={2024},
  url={https://arxiv.org/abs/2306.13649}
}

@misc{lu2025onpolicy,
  title={On-Policy Distillation},
  author={Lu, Kevin},
  year={2025},
  howpublished={Thinking Machines Lab blog, \url{https://thinkingmachines.ai/blog/on-policy-distillation/}},
  note={Accessed 2026-09-06}
}

@inproceedings{hendrycks2021math,
  title={Measuring Mathematical Problem Solving With the {MATH} Dataset},
  author={Hendrycks, Dan and Burns, Collin and Kadavath, Saurav and Arora, Akul and Basart, Steven and Tang, Eric and Song, Dawn and Steinhardt, Jacob},
  booktitle={Advances in Neural Information Processing Systems},
  year={2021},
  url={https://arxiv.org/abs/2103.03874}
}
